\documentclass[conference]{IEEEtran}
\ifCLASSINFOpdf
\else
\fi
\title{Bare Demo of IEEEtran.cls\\ for IEEE Conferences}

\author{\IEEEauthorblockN{Michael Shell}
\IEEEauthorblockA{School of Electrical and\\Computer Engineering\\
Georgia Institute of Technology\\
Atlanta, Georgia 30332--0250\\
Email: http://www.michaelshell.org/contact.html}
\and
\IEEEauthorblockN{Homer Simpson}
\IEEEauthorblockA{Twentieth Century Fox\\
Springfield, USA\\
Email: homer@thesimpsons.com}
\and
\IEEEauthorblockN{James Kirk\\ and Montgomery Scott}
\IEEEauthorblockA{Starfleet Academy\\
San Francisco, California 96678--2391\\
Telephone: (800) 555--1212\\
Fax: (888) 555--1212}}

\usepackage[hyphens]{url}  
\usepackage{algorithm}
\usepackage{algorithmic}

\usepackage{amsfonts}       
\usepackage{xcolor}         

\usepackage{graphicx}
\usepackage{amsmath}
\usepackage{booktabs}
\usepackage{algorithm}
\usepackage{algorithmic}
\usepackage{multirow}
\usepackage{amsthm}
\usepackage{amssymb}
\usepackage{mathrsfs}
\usepackage{subfigure}

\usepackage{newfloat}
\usepackage{listings}

\title{FedMR: Fedreated Learning via Model Recombination}

\author{
  Ming Hu,
 Zhihao Yue,
 Zhiwei Ling,
 Xian Wei,
 Mingsong Chen$^*$\\
 Shanghai Key Lab of Trustworthy Computing, East China Normal University\\
$^*$Corresponding Author, Email: mschen@sei.ecnu.edu.cn
}

\begin{document}

\maketitle

\begin{abstract}

As a promising  privacy-preserving 
machine learning method, Federated Learning (FL)  
enables global model training across  clients
without compromising their  confidential local
data. However, existing FL methods suffer from the 
 problem of  low  inference performance for unevenly distributed data, 
 since most of them rely on Federated Averaging 
 (FedAvg)-based aggregation.
By  averaging  model parameters in a coarse manner, 
FedAvg eclipses the individual characteristics 
of local models, 
which strongly limits the inference capability of FL. 
Worse still, in each round of  FL training, FedAvg dispatches  
the same initial local models to clients, which can easily 
result in {\it stuck-at-local-search}  for optimal global models.
%
To address the above issues, 
this paper proposes a novel and effective
FL paradigm 
named FedMR (Federating Model Recombination).  
Unlike conventional FedAvg-based methods, the cloud server  of 
FedMR shuffles each layer of collected local models and recombines them to achieve new  models for local training on clients. Due to the fine-grained model  recombination and  local training
in each FL round, FedMR can quickly figure out 
one globally optimal  model for all the clients. 
Comprehensive experimental results 
demonstrate that, compared with state-of-the-art FL methods, 
FedMR can significantly improve the 
inference  accuracy without causing extra communication overhead.
\end{abstract}

\IEEEpeerreviewmaketitle

\section{Introduction}

Along with the increasing popularity of 
Artificial Intelligence (AI), Federated Learning (FL) \cite{fedavg}
has been widely acknowledged as a promising means to  design  
 large-scale distributed  AI applications, e.g., 
 AI Internet of Things (AIoT) applications~\cite{tcad_zhang_2021,iot_haya_2022}, 
healthcare systems~\cite{cvpr_quande_2021,kdd_qian_2021}, 
and recommender  systems~\cite{icassp_david_2019}, where the data on 
clients are assumed to be secure and cannot be 
accessed by other clients.  Unlike conventional  Deep Learning (DL)-based  methods,
FL supports  the  collaborative  training of a global DL model  across clients without compromising their  data privacy \cite{nips_naman_2021,aaai_bo_2022,aaai_chendi_2022}.
In each  FL training round, the cloud server firstly dispatches an intermediate 
model to its selected clients for  local training   
and then gathers the corresponding gradients of trained 
models from clients for aggregation.
In this way, only clients can touch their own data, thus 
the client  privacy  can be guaranteed.

Based on the cloud-client architecture, FL enables effective
collaborative  training towards a global optimal model 
for all the involved  clients. However, when dealing with the 
unevenly  distributed  data  on  clients, 
existing main-stream FL methods  suffer from the notorious
 problem of ``weight divergence''~\cite{kairouz2021advances}.
 Especially when the  data on clients are non-identically and independently distributed (Non-IID) \cite{infocom_hao_2020,iclr_2021_durmus}, the convergence directions
 of local models on the devices and the aggregated global model on the cloud
 are significantly
 inconsistent, thus inevitably
 resulting in the inference performance degradation and slow convergence of global models. 
 Aiming  to  mitigate such a  phenomenon and  improve FL performance  in non-IID scenarios, various  FL methods have been studied, 
 e.g., client grouping-based methods~\cite{axiv_ming_2021}, global 
 control variable-based methods~\cite{icml_scaffold,aaai_yutao_2021,fedprox}, 
 and Knowledge Distillation (KD)-based methods\cite{nips_tao_2020,icml_zhuangdi_2021}. 
 The basic ideas of these solutions are to 
 guide the local training on clients~\cite{icml_scaffold,aaai_yutao_2021} or 
 adjust parameters
 weights 
 for model aggregation~\cite{axiv_ming_2021,nips_tao_2020,icml_zhuangdi_2021}.
 

Although more and more  FL  methods  are proposed to 
alleviate the  impact of data heterogeneity, most of them 
adopt the well-known  Federated Averaging
(FedAvg)-based aggregation, which strongly limits their  inference 
capability. This is mainly because    FedAvg operations conducted by the cloud server
only   average  the parameters of collected local models in a coarse manner, 
where the specific characteristics of client data  
learned  by  local models are almost neglected. 
As a result, an intermediate
global model based on 
simple statistical averaging cannot accurately
reflect both the individual 
efforts and potential of local models in searching for  global optimal models. 
Worse still, after FedAvg the cloud server will dispatch the same 
global model to the selected clients for the next round of local training. 
Due to the same initial models for local training, 
existing FL methods can easily 
get {\it stuck-at-local-search} for optimal global models, thus further 
degrading the overall inference performance (i.e., classification accuracy and convergence rate) of obtained global models. 
 {\it Clearly, how to  break through the prevalent FedAvg-based aggregation 
 paradigm to  achieve better inference performance is 
 becoming an urgent issue 
in the design of modern FL applications.}



To address the above challenge, this paper presents  a novel FL paradigm named 
FedMR based on our proposed Federated Model Recombination method. 
Instead of using FedAvg for  model aggregation,  in each FL training round, 
FedMR decouples each local model into multiple layers and recombines them to form 
new local  models. Based on a new round of local training on clients, the 
dependencies between the layers of recombined models will be reconstructed. 
In this way, 
since the decomposed layers consist of partial  
knowledge learned by previous
local models and each FL training round  adopts different
 models for local training
 on different clients, FedMR can effectively 
escape from the stuck-at-local-search and guide the 
evolution of local models towards global optimal models, 
thus improving the overall FL performance. 
This paper makes the following three major contributions:
\begin{itemize}
    \item
 We present a novel FL paradigm named FedMR and its  
 implementation details, which enable  fine-grained model recombination
 and local training to achieve better FL performance than traditional 
 FedAvg-based methods.
 

  
\item 
We introduce a  two-stage  training scheme
for FedMR, which not only accounts for the individual efforts of
local models during the training but also combines the merits of 
FedAvg-based aggregration and model recombination to accelerate the 
overall FL traning process. 

\item 
We conduct comprehensive experiments using various models and datasets
to evaluate the  effectiveness and pervasiveness
of  FedMR  in both 
IID and non-IID scenarios. 
\end{itemize}


\section{Preliminaries  and Related Work}\label{sec:relatedwork}

\subsection{Preliminaries  to Federated Learning}

Typically, FL adopts the cloud-based architecture to enable the 
aggregation of local models. To achieve  a global model that best fits for all the clients,   
 each FL training round typically consists of three step: i)
{\it model dispatching} step where the cloud server dispatches an
intermediate 
global model to multiple clients; ii) the {\it local training step} that 
trains
local models based on the local data of clients;
and iii) the {\it aggregation step} where   the cloud server aggregates all the uploaded gradients to update the original global model. 
So far, almost FL methods aggregate local models based on 
 FedAvg \cite{fedavg} defined as follows:
\begin{footnotesize}
\begin{equation}
\begin{split}
&\min_{w} F(w) = \frac{1}{N}\sum_{i = 1}^{N} f_i(w), \\
\ s.t., &\ f_i(w) = \frac{1}{n_i} \sum_{j = 1}^{n_i} \ell (w;x_j;y_j),
\end{split}
\end{equation}
\end{footnotesize}
where $N$ is the total number of clients, $n_i$ is the number of data samples hosted by the $i^{th}$ client, $\ell$ denotes the customer-defined loss function (e.g., cross-entropy loss), $x_j$ denotes a sample, and $y_j$ is the label of $x_j$.

\begin{figure*}[ht] 
	\begin{center} 
		\includegraphics[width=0.76\textwidth]{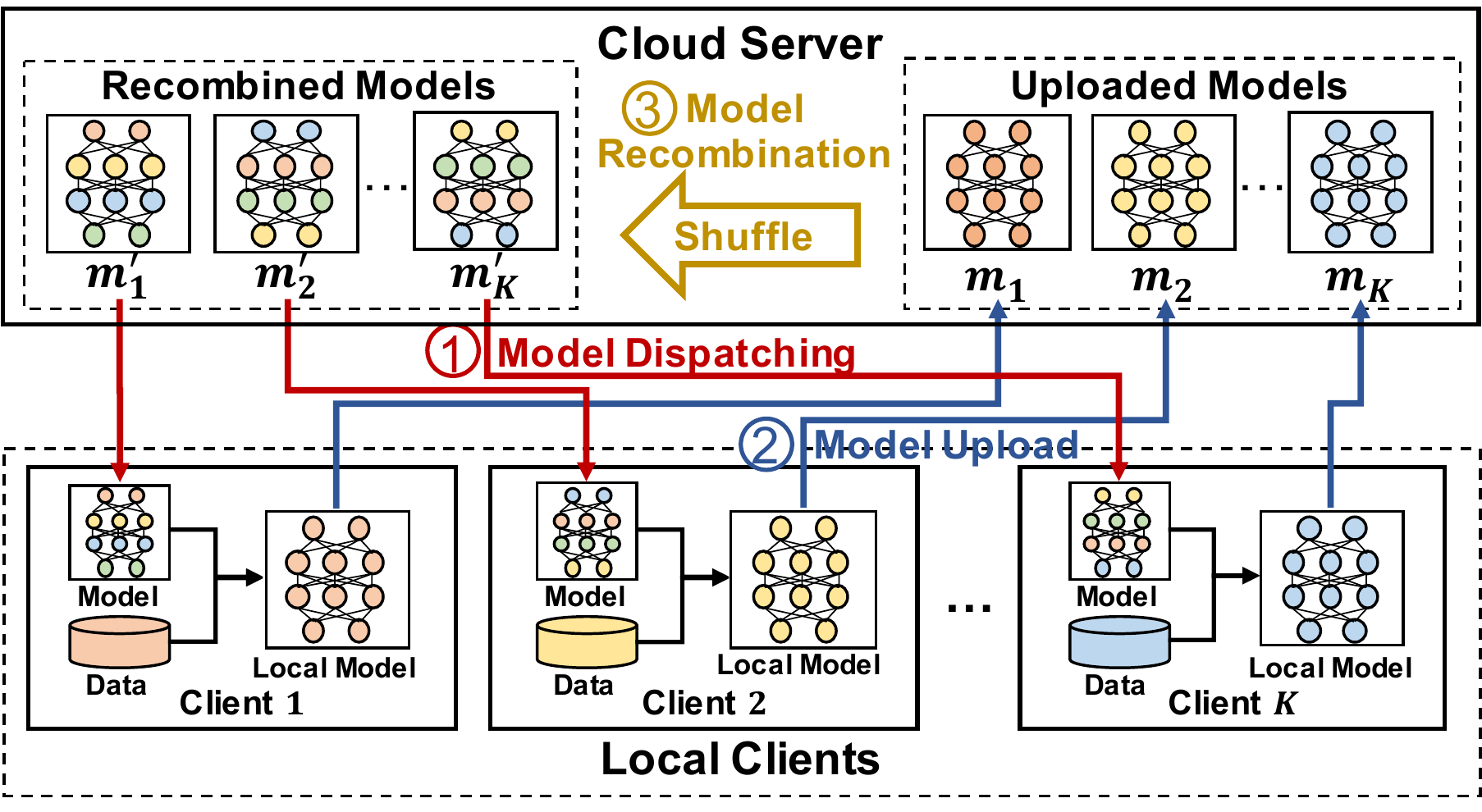}
		\caption{Framework and workflow of FedMR.}
		\label{fig:framework} 
	\end{center}
\end{figure*}

\subsection{Performance  Optimization Methods for FL}

Although FL  enables distributed training of a global model
for clients without compromising their data privacy, it 
still suffers from the problem of uneven data distributions on clients. 
To address this problem,  numerous
solutions  have been proposed to improve the performance of FL , which can be classified into three categories, i.e., client grouping-based methods, global control variable-based methods, and knowledge distillation-based methods.

The {\it device grouping-based methods}
group and select clients for aggregation based on the data   similarity between clients. For example, FedCluster~\cite{bigdata_cheng_2020} divides
clients into multiple clusters and performs multiple cycles of  meta-update
to boost the overall FL convergence. Based on either 
sample size or model similarity, 
CluSamp~\cite{icml_yann_2021} groups clients to achieve a better 
 client representativity and a reduced variance of client stochastic 
 aggregation parameters in FL.
By modifying the  penalty terms of loss functions during  FL 
training, the {\it global control variable-based methods}
can be used to 
smooth the FL convergence process. 
For example, 
FedProx~\cite{fedprox} regularizes local loss functions with the squared  distance between local models and the global model, which can stabilize the model convergence. Similarly, SCAFFOLD~\cite{icml_scaffold}  uses 
global control variables to correct the ``client-drift'' problem in the local training process. By inferring the composition of training data in each FL round, Wang et al.~\cite{aaai_lixu_2021} presented a novel FL architecture to alleviate the impact of the data imbalance issue.
As a promising FL  optimization technology, {\it Knowledge Distillation (KD)-based methods} adopt soft targets generated by the ``teacher model'' to guide the training of ``student models''.
For example, by leveraging a proxy dataset, Zhu et al.~(\cite{icml_zhuangdi_2021}) proposed a data-free knowledge distillation method named FedGen to address the heterogeneous FL problem using a built-in generator. With ensemble distillation, FedDF~\cite{nips_tao_2020} accelerates the FL training by training 
 the global  model through unlabeled data on the outputs of local models. 
Based on transfer learning, FedMD~\cite{axiv_daliang_2019} trains models on both public datasets and private datasets to mitigate the data heterogeneity.

Although the above  methods can effectively optimize
FL   performance from different perspectives,
most of them cannot avoid 
non-negligible communication and computation overheads or  the risk of data privacy exposure. Notably, all the above methods are based on FedAvg, which 
only conducts coarse-grained model aggregation. Consequently, the 
inference capabilities
of local models are strongly restricted. 
To the best of our knowledge,  FedMR is the first attempt  that uses 
 model recombination and different models
for fine-grained FL. Since FedMR considers the
specific characteristics and efforts of local models, 
it can further mitigate the  weight divergence problem, thus achieving 
better inference performance than state-of-the-art   FL methods.


\section{Our FedMR Approach}

\subsection{Overview of FedMR}


FedMR adopts the same cloud-based
architecture as the one used by  conventional FL methods, where the cloud server conducts  model recombination operations and  clients perform  the local training.   Figure~\ref{fig:framework} presents the framework and workflow of FedMR, where each FedMR training round involves three
specific steps  as follows:
\begin{itemize}
    \item \textbf{Step 1 (Model Dispatching):} The cloud server dispatches $K$ recombined models to 
    $K$ selected clients according to their indices, where $K$ is the number of clients that participate in the FL
    training. Note that, unlike traditional FedAvg-based FL methods, 
    in FedMR different clients will receive different local models for the local training purpose.
    \item \textbf{Step 2 (Model Upload):} 
    Once the local training is finished, a client needs to upload the parameters of its  trained local model to the cloud server.
    \item \textbf{Step 3 (Model Recombination):} After collecting all the  local model information
    in the FL training round,
    the cloud server firstly decomposes the   
    local  models  into multiple layers individually
    in the same manner, and then conducts the random shuffling of the same layers among different local models. Finally, by concatenating layers
    from different sources in order,  a new local model can be 
    reconstructed by the cloud server. Note that any decomposed layer of the uploaded model will be eventually used by one and only one of the 
    recombined models. 
\end{itemize}

\subsection{FedMR Implementation}

Algorithm~\ref{alg:fedmr} presents 
the  implementation details  of FedMR.
 Line~\ref{line:init} initializes the model list $L_m$, which includes $K$ initial models. Lines~\ref{line:trainStart}-\ref{line:trainEnd} performs  
$rnd$ rounds of  FedMR  training.
In each  round,  Line~\ref{line:clientSel}  selects $K$ random clients to participate the model training and creates a client list  $L_r$.
Lines 4-7 conduct the local training on clients in parallel, 
where Line 5 applies the local model $L_m[i]$ on client $L_r[i]$ for local training by using the
function {\it ClientUpdate},
and Line 6 achieves a new local model after the local training. 
After the cloud server receives 
all the  $K$ local models,  Line~\ref{line:modelRecomb} uses the 
function {\it ModelRcombine}  to recombine local models and 
generate $K$ new local models, 
which are saved in $L_m$ as shown in Line 9.
Finally, Line 11 generates 
an optimal global model based on the 
local models in $L_m$, which will be 
reported by Line 12. 
The following parts will give the detailed implementation 
of the functions used in  FedMR.

\begin{algorithm}[ht]
\caption{Implementation of  FedMR}
\label{alg:fedmr}
\textbf{Input}:\\
i) $rnd$, \# of training rounds; \\
ii) $S_{c}$, the set of clients; \\
iii) $K$, \# of clients  participating in each FL round.\\
\textbf{Output}: \\
i) $m_{glb}$, the trained global model\\
\textbf{FedMR}($rnd$,$S_{dev}$,$K$)
\begin{algorithmic}[1] 
\STATE $L_m\leftarrow [w^1_1, w^2_1,...,w^{K}_1]\ \ \ $ // initialize model list  \;\label{line:init}
\FOR{$r$ = 1, ..., $rnd$}\label{line:trainStart}
\STATE $L_r\leftarrow$ Random select $K$ clients from $S_{c}$\;\label{line:clientSel}
\\ /*parallel for block*/
\FOR{$i$ = 1, ..., $K$}   
\STATE $v_{r+1}^i\leftarrow ${\it ClientUpdate}$(L_m[i],L_r[i])$\;\label{line:clientUpdate}
\STATE $L_m[i]\leftarrow v_{r+1}^i$\;
\ENDFOR
\STATE $[w_{r+1}^1,w_{r+1}^2,..., w_{r+1}^K]\leftarrow ${\it ModelRcombine}$(L_m)$\label{line:modelRecomb}\;
\STATE $L_m\leftarrow [w_{r+1}^1,w_{r+1}^2,..., w_{r+1}^K]$\;
\ENDFOR\label{line:trainEnd}
\STATE $m^{glb}\leftarrow ${\it GlobalModelGen}$(L_m)$\label{line:modelAggr}\;
\STATE \textbf{return} $m_{glb}$\;
\end{algorithmic}
\end{algorithm}

\subsubsection{Local Model Training (ClientUpdate).} 
Unlike conventional FL methods that 
conduct local  
training on clients starting from the same 
aggregated  model, in each training round FedMR uses
different recombined models (i.e., $K$ models in the 
model list $L_m$) for the local training purpose. 
Note that, in the whole training phase,
FedMR  only uses  $K$ ($K\leq |S_c|$) models, since there are only $K$ devices selected in each  training round. 
Let $w^c_r$ be the parameters of some
model that is dispatched 
to the $c^{th}$ client in the $r^{th}$ training round. In the $r^{th}$ training round, we dispatch 
the $i^{th}$ model in $L_m$  to its corresponding client as follows:
\begin{equation}
w^{L_r[i]}_{r} = L_m[i].
\end{equation}
Based on the obtained recombined model, FedMR conducts the local training on client $L_r[i]$ as follows:
\begin{equation}
\begin{split}
v^{L_r[i]}_{r+1} &=w^{L_r[i]}_{r} - \eta \nabla f_{L_r[i]}(w^{L_r[i]}_{r}),\\
s.t. \ f_{L_r[i]}(w^{L_r[i]}_{r}) &= \frac{1}{|D_{L_r[i]}|} \sum_{j = 1}^{|D_{L_r[i]}|} \ell (w^{L_r[i]}_{r};x_j;y_j),\\
\end{split}
\end{equation}
where $v^{L_r[i]}_{r}$ indicates parameters of the trained local model, $D_{L_r[i]}$ denotes
the dataset of client $L_r[i]$, 
$\eta$ is the  learning rate, $\ell()$ is the loss function, $x_j$ is the $j^{th}$ sample in $D_{L_r[i]}$, and $y_j$ is the label of $x_j$.
Once the local training is finished , 
 the  client needs to upload the parameters of 
 its trained 
  local model  to the cloud server by updating 
  $L_m$ using 
\begin{equation}
L_m[i] = v^{L_r[i]}_{r+1}.
\end{equation}
Note that, similar to traditional FL methods, 
in each training round FedMR needs to transmit 
the parameters of $2K$ models between the cloud server and its selected clients.

\subsubsection{Model Recombination (ModelRecombine).}
Typically, a DL model consists of  multiple layers, e.g., convolutional layers, pooling layers, Fully Connected (FC) layers. In our model recombination method, we  do not take 
 the pooling 
 layers into account, since they are constant. To simplify the 
description of our model recombination method, we
do not explicitly present the layer types here. 
Let $w_x=\{l_1^x,l_2^x,...,l_n^x\}$ be
the parameters of  model $x$, where $l_i^x$ ($i\in [1,n]$) denotes the  parameters of the $i^{th}$ layer of model 
$x$. 

\begin{figure}[ht] 
	\begin{center} 
		\includegraphics[width=0.5\textwidth]{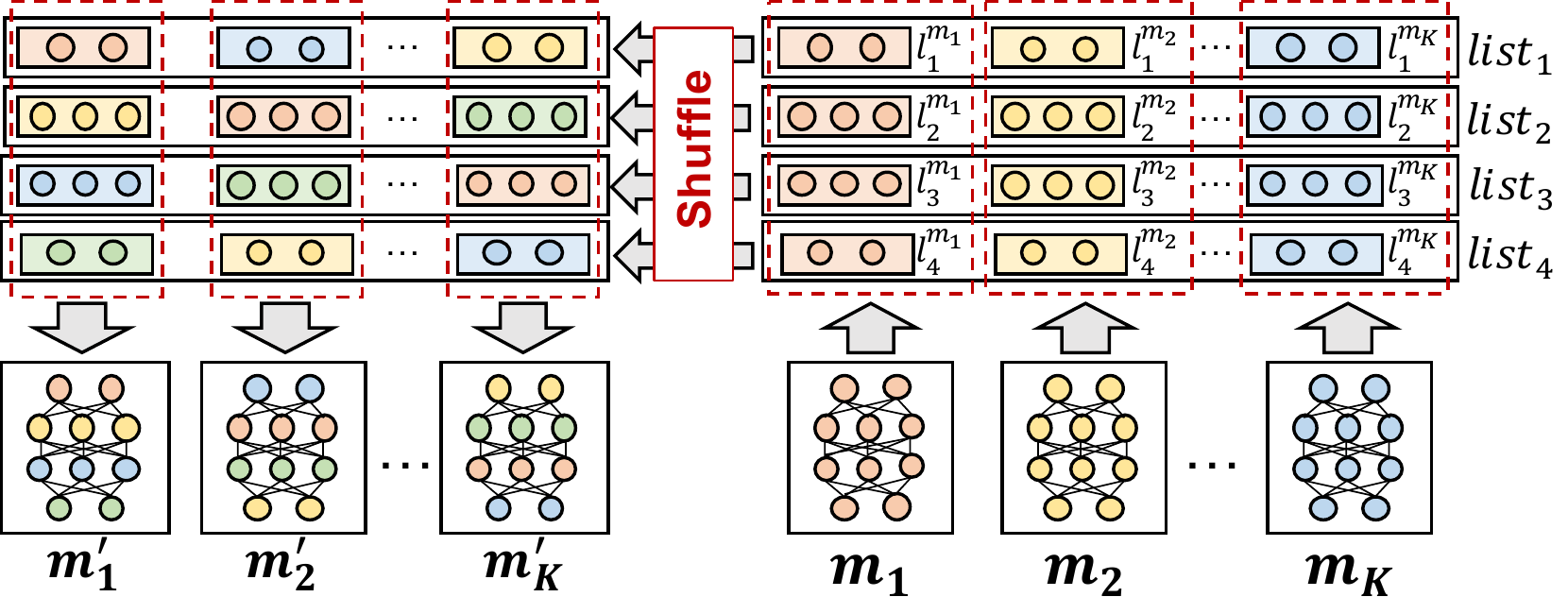}
		\caption{An example of model recombination.}
		\label{fig:mr} 
	\end{center}
\end{figure} 

In each FL round,  FedMR needs to conduct the model recombination base on $L_m$ to obtain new models for the local training. 
Figure~\ref{fig:mr} shows an example of model recombination based on the shuffling of model layers.
When receiving all  the trained local models (i.e., $m_1$,$m_2$,...,$m_K$) from clients, 
firstly the cloud server needs to decouple
the layers of these models individually.
For example, the model $m_1$ can be decomposed into 
four layers. Assuming that the local models are with an architecture of $w$, 
to enable the recombination, FedMR then constructs
$n$ lists, where the $k^{th}$ ($k\in [1,n]$) list 
contains
all the $k^{th}$ layers of the models in $L_m$. As an example shown in Figure~\ref{fig:mr}, 
FedMR constructs four lists (i.e., $list_1$-$list_4$)
for the $K$ models (i.e., $m_1$-$m_K$), where each list consists of $K$ elements (i.e.,  $K$ layers with the same index). 
Next, FedMR  
shuffles the elements within each list, and   generates $|L_m|$ recombined models based on shuffled results.  For example, the top three layers of 
the recombined model 
$m^\prime_1$ come from the models   $m_1$, $m_2$
     and $m_K$, respectively.

\subsubsection{Global Model Generation (GlobalModelGen).}

Although there are only $K$ models involved in 
each local training round, 
the goal of FedMR is to obtain a global model that can accommodate all the $|S_c|$ clients. 
When the local training finishes, in  FedMR
the cloud server will collect all the parameters of local models and conduct FedAvg-style aggregation
on them to achieve the expected global model as follows:
\begin{equation}
\begin{split}
w^{glb} = \frac{1}{K}\sum_{i = 1}^{K} w_{rnd+1}^{i}
\end{split}
\end{equation}
where $w_{rnd+1}^i$ is parameters of the $i^{th}$  model in $L_m$ when the  whole local training completes. Note that the global model will be dispatched by the cloud server to all
the clients for the  purpose of inference
rather than local training.


\subsection{Two-Stage Training Scheme for FedMR}
Traditional FedAvg-based  methods  conduct
coarse-grained model aggregation by periodically 
learning local knowledge from clients using the same  
models. Unlike these methods,  FedMR can learn the knowledge  from all the global data based on our proposed model recombination 
method. For  model recombination, FedMR  needs to break the 
dependencies between layers of each local model first, 
and then  reconstruct the new dependencies between newly 
shuffled layers  for different recombined models via 
respective local training. 
Based on  continuous iterations of  
the coupling  of  model recombination and 
local training, each layer of a local model will gradually 
adapt to all the data across clients. Since FedMR conducts FL based on
the recombination of model layers, it can be considered as a 
 fine-grained FL paradigm.

Although  FedMR enables finer
 FL training, when  starting from blank models,  
 FedMR converges more slowly than 
traditional FL methods at the beginning. This is mainly because, due to the low matching degree between  layers in the recombined models,
the model recombination operation in this stage 
requires more local training time to re-construct the new dependencies between layers. To accelerate the overall convergence, we 
propose a  two-stage training scheme for FedMR, consisting of both the  {\it aggregation-based pre-training stage} and  {\it model recombination stage}.
 In the first stage, we 
train the local models coarsely  using the 
 FedAvg-based aggregation, which can quickly form a pre-trained 
 global model. In the second stage, starting from the
 pre-trained models, FedMR dispatches recombined models to clients
 for local training. Due to the synergy 
  of both FL paradigms, the overall FedMR training time can be reduced.

\section{Experimental Results}
\subsection{Experimental Settings}
To evaluate the effectiveness of FedMR, we implemented FedMR
on top of a  cloud-based
architecture. Since it is impractical to allow 
all the clients to get involved in the training processes simultaneously, 
we assumed that there are only 10\% of clients participating the local training 
in each FL round. To enable fair comparison, all the investigated 
FL methods including FedMR set their
SGD optimizer  with a learning rate of 0.01 and a momentum of 0.9.
For each client,  we set the batch size of local training  to 50,  and performed 
five epochs for each local training. All the experimental results were obtained 
from an Ubuntu workstation with Intel i9 CPU, 32GB memory, and NVIDIA  RTX 3080 GPU.

\begin{table*}[ht]
\centering
\caption{Test accuracy comparison for both non-IID and  IID scenarios using three DL models}
\label{tab:acc}
\scriptsize
\begin{tabular}{|c|c|c|c|c|c|c|c|c|c|}
\hline
\multirow{2}{*}{Model} & \multirow{2}{*}{Dataset} & Heteroge. & \multicolumn{7}{c|}{Test Accuracy (\%)} \\
\cline{4-10}
 & & Settings & FedAvg & FedProx &  SCAFFOLD & MOON & FedGen & CluSamp & {\bf FedMR} \\
\hline
\hline
\multirow{9}{*}{CNN} & \multirow{4}{*}{CIFAR-10} & $\alpha=0.1$& $46.12\pm 2.35$& $47.17\pm 1.65$& $49.12\pm 0.91$ & $42.61\pm 2.65$ & $ 49.27\pm 0.85 $ & $47.09\pm 0.97$ & ${\bf 54.22\pm 1.25}$\\
                      &  & $\alpha=0.5$ & $52.82\pm 0.91$ & $53.59\pm 0.88$ & $54.50\pm 0.44$ & $53.56\pm 1.74$ & $51.77\pm 0.73$  & $54.00\pm 0.38$ & ${\bf 59.13\pm 0.65}$\\
                      &  & $\alpha=1.0$ & $54.78\pm 0.56$ & $54.96\pm 0.60$ & $56.75\pm 0.26$ & $54.51\pm 1.24$ & $55.38\pm 0.66$  & $55.82\pm 0.73$ & ${\bf 61.10\pm 0.49}$\\
                      &  & $IID$ & $57.64\pm 0.22$ & $58.34\pm 0.15$ & $59.98\pm 0.22$ & $57.33\pm 0.30$  & $58.71\pm 0.19$ & $57.32\pm 0.21$ & ${\bf 62.07\pm 0.29}$\\
\cline{2-10}
& \multirow{4}{*}{CIFAR-100} & $\alpha=0.1$ & $28.37\pm 1.10$ & $28.11\pm 1.03$ & $30.32\pm 1.05$ & $28.15\pm 1.54$ & $28.18\pm 0.58$  & $28.63\pm 0.63$   & ${\bf 33.33\pm 0.87}$\\
                     &  & $\alpha=0.5$     & $30.01\pm 0.56$ & $32.16\pm 0.50$ & $33.49\pm 0.73$ & $30.93\pm 0.49$ & $29.55\pm 0.41$  & $33.04\pm 0.41$ & ${\bf 36.96\pm 0.30}$\\
                     &  & $\alpha=1.0$      & $32.34\pm 0.65$ & $32.78\pm 0.13$ & $34.95 \pm 0.58$ & $31.46\pm 0.66$ & $31.88\pm 0.65$  & $32.92\pm 0.31$ & ${\bf 38.05\pm 0.24}$\\
                     &  & $IID$ & $32.98\pm 0.20$ & $33.39\pm 0.25$ & $35.11\pm 0.23$ & $32.39\pm 0.19$  & $32.43\pm 0.20$ & $34.97\pm 0.24$ & ${\bf 40.01\pm 0.11}$\\
\cline{2-10}
& \multirow{1}{*}{FEMNIST} & $-$ & $81.67\pm 0.36$ & $82.10\pm 0.61$ & $81.65\pm 0.21$ & $81.13\pm 0.39$ & $81.95 \pm 0.36$  & $80.80\pm 0.40$  &  ${\bf 82.73\pm 0.36}$ \\
\hline

\hline
\multirow{10}{*}{ResNet-20} & \multirow{4}{*}{CIFAR-10} & $\alpha=0.1$& $45.11\pm 2.13$& $45.45\pm 3.42$& $50.46\pm 1.76$ & $46.38\pm 2.66$ & $42.71\pm 3.48$  & $44.87\pm 1.65$ & ${\bf 62.09\pm 1.77}$\\
                      &  & $\alpha=0.5$ & $60.56\pm 0.95$ & $59.52\pm 0.74$ & $58.85\pm 0.85$ & $60.47\pm 0.68$ & $60.29\pm 0.68$  & $59.55\pm 1.00$ & ${\bf 74.00\pm 0.32}$\\
                      &  & $\alpha=1.0$ & $62.99\pm 0.62$ & $61.47\pm 0.66$ & $61.63\pm 0.78$ & $61.99\pm 0.68$ & $63.81\pm 0.33$  & $63.32\pm 0.71$ & ${\bf 76.92\pm 0.38}$\\
                      &  & $IID$ & $67.12\pm 0.27$ & $66.06\pm 0.22$ & $65.20\pm 0.27$  & $66.19\pm 0.22$ & $65.89\pm 0.17$ & $65.62\pm 0.23$ & ${\bf 77.94\pm 0.14}$\\
\cline{2-10}
& \multirow{4}{*}{CIFAR-100} & $\alpha=0.1$ & $31.90\pm 1.16$ & $33.00\pm 1.21$ & $35.71\pm 0.62$ & $32.91\pm 0.70$ & $32.40\pm 1.45$  & $34.34\pm 0.52$   & ${\bf 45.13\pm 1.05}$\\
                     &  & $\alpha=0.5$     & $42.45\pm 0.53$ & $42.83\pm 0.54$ & $42.33\pm 1.23$ & $41.76\pm 0.22$  & $42.72\pm 0.32$ & $42.07\pm 0.39$ & ${\bf 54.73\pm 0.27}$\\
                     &  & $\alpha=1.0$      & $44.22\pm 0.36$ & $44.35\pm 0.36$ & $43.28\pm 0.61$ & $42.92\pm 0.67$ & $44.75\pm 0.57$  & $43.29\pm 0.41$ & ${\bf 56.96\pm 0.31}$\\
                     &  & $IID$ & $44.42\pm 0.18$ & $45.16\pm 0.24$ & $44.37\pm 0.19$ & $46.13\pm 0.13$  & $45.21\pm 0.19$ & $43.59\pm 0.24$ & ${\bf 59.25\pm 0.35}$\\
\cline{2-10}
& \multirow{1}{*}{FEMNIST} & $-$ & $78.47\pm 0.40$ & $79.74\pm 0.54$ & $76.14\pm 0.90$ & $79.50\pm 0.46$ & $79.56\pm 0.34$  & $79.28\pm 0.42$  &  ${\bf 81.27\pm 0.31}$ \\
\hline

\hline
\multirow{9}{*}{VGG-16} & \multirow{4}{*}{CIFAR-10} & $\alpha=0.1$& $63.79\pm 3.90$& $63.35\pm 4.31$& $64.18\pm 3.86$ & $60.19\pm 3.73$ & $66.52\pm 1.46$  & $66.91\pm 1.83$ & ${\bf 74.38\pm 0.71}$\\
                      &  & $\alpha=0.5$ & $78.14\pm 0.67$ & $77.70\pm 0.45$ & $76.22\pm 1.37$ & $77.41\pm 0.77$ & $78.9\pm 0.39$  & $78.82\pm 0.40$ & ${\bf 82.86\pm 0.37}$\\
                      &  & $\alpha=1.0$ & $78.55\pm 0.21$ & $79.10\pm 0.28$ & $76.99\pm 1.01$ & $78.81\pm 0.41$  & $79.75\pm 0.26$ & $80.00\pm 0.37$ & ${\bf 84.45\pm 0.23}$\\
                      &  & $IID$ & $80.02\pm 0.05$ & $80.77\pm 0.22$ & $78.80\pm 0.07$ & $81.11\pm 0.12$ & $80.00\pm 0.27$  & $80.96\pm 0.12$ & ${\bf 85.87\pm 0.23}$\\
\cline{2-10}
& \multirow{4}{*}{CIFAR-100} & $\alpha=0.1$ & $46.60\pm 1.45$ & $45.88\pm 3.35$ & $45.79\pm 1.77$ & $42.74\pm 1.10$ & $49.04\pm 0.63$  & $48.04\pm 1.76$   & ${\bf 56.60\pm 0.83}$\\
                     &  & $\alpha=0.5$     & $55.86\pm 0.64$ & $55.79\pm 0.56$ & $55.30\pm 0.61$ & $53.29\pm 0.79$ & $56.40\pm 0.37$  & $56.23\pm 0.34$ & ${\bf 65.04\pm 0.16}$\\
                     &  & $\alpha=1.0$      & $57.55\pm 0.51$ & $57.40\pm 0.32$ & $55.43\pm 0.45$ & $54.67\pm0.55$ & $57.15\pm 0.27$  & $57.95\pm 0.35$ & ${\bf 66.28\pm 0.34}$\\
                     &  & $IID$ & $58.30\pm 0.23$ & $58.49\pm 0.11$ & $56.51\pm 0.08$ & $57.39\pm 0.24$  & $57.62\pm 0.18$ & $58.14\pm 0.20$ & ${\bf 66.28\pm 0.11}$\\
\cline{2-10}
& \multirow{1}{*}{FEMNIST} & $-$ & $84.22\pm 0.46$ & $83.98\pm 0.48$ & $82.65\pm 0.74$ & $79.09\pm0.42$  & $84.69\pm0.28$ & $84.32\pm 0.36$  &  ${\bf 85.36\pm 0.21}$\\
\hline
\end{tabular}
\end{table*}

\subsubsection{Baseline Method Settings} We compared the test accuracy of FedMR with six baseline  methods, 
 i.e.,  FedAvg~\cite{fedavg}, FedProx~\cite{fedprox}, SCAFFOLD~\cite{icml_scaffold}, MOON~\cite{moon}, FedGen~\cite{icml_zhuangdi_2021}, and CluSamp~\cite{icml_yann_2021}. 
 Here,   FedAvg is the most classical FL method, while 
 the other five methods are the state-of-the-art (SOTA) 
representatives of the three kinds of FL  
optimization methods introduced in the related work section.
Specifically, FedProx, SCAFFOLD, and MOON are global control variable-based methods, FedGen is a KD-based approach, and CluSamp is a device grouping-based method.
For FedProx, we used a hyper-parameter $\mu$  to control the weight of its proximal term, where the best values of $\mu$  for CIFAR-10, CIFAR-100, and FEMNIST
are 0.01, 0.001, and 0.1, respectively.
For FedGen, we adopted  the same sever settings  presented in \cite{icml_zhuangdi_2021}.
For CluSamp, the clients were clustered based on the model gradient similarity
as described in \cite{icml_yann_2021}.



\subsubsection{Dataset  Settings}
We investigated the performance of our approach on 
three well-known datasets, i.e., CIFAR-10,  CIFAR-100 \cite{data}, 
and FMNIST~\cite{leaf}. 
We adopted
the Dirichlet distribution~\cite{measuring} to control the heterogeneity of 
client data for both CIFAR-10  and CIFAR-100.
Here, the notation $Dir(\alpha)$ indicates a different Dirichlet distribution
controlled by $\alpha$, where a smaller $\alpha$ means higher data heterogeneity
of clients. Note that, different from datasets CIFAR-10 and CIFAR-100, 
the raw data of FEMNIST are naturally non-IID distributed. Since FEMNIST takes 
various kinds of imbalances (e.g., data heterogeneity, data imbalance and class imbalance) into account, we did not apply the 
Dirichlet distribution on FEMNIST. 
For both CIFAR-10 and CIFAR-100, we assumed that there are 100 clients in total participating
in FL. For FEMNIST,  we only considered one non-IID scenario 
involving 180 clients, where each client hosts more than 100 local
data samples\footnote{The  settings were obtained from  LEAF benchmark by using the  
command: ./preprocess.sh -s niid –sf 0.05 -k 100 -t sample.}.

\subsubsection{Model  Settings}
To demonstrate the pervasiveness of our approach, we 
developed different FedMR implementations
based on 
three different  DL models (i.e., CNN, ResNet-20, VGG-16). 
Here, we obtained the CNN model from \cite{fedavg}, which 
consists of 
two convolutional layers and  two FC layers. 
When conducting 
FedMR based on the CNN model, we directly applied
the model recombination for local training on 
it without pre-training a global model, since CNN here  
only has four layers. 
We obtained both ResNet-20 and VGG-16 models from Torchvision \cite{models}. 
When performing
FedMR based on ResNet-20 and VGG-16, 
due to the deep structure of both models, 
we adopted the two-stage training scheme, where the first stage 
lasts for 100 rounds based on model aggregation to obtain a pre-trained global model.


\begin{figure*}[ht]
	\centering
	\subfigure[CIFAR-100 with $\alpha=0.1$]{
		\centering
		\includegraphics[width=0.25\textwidth]{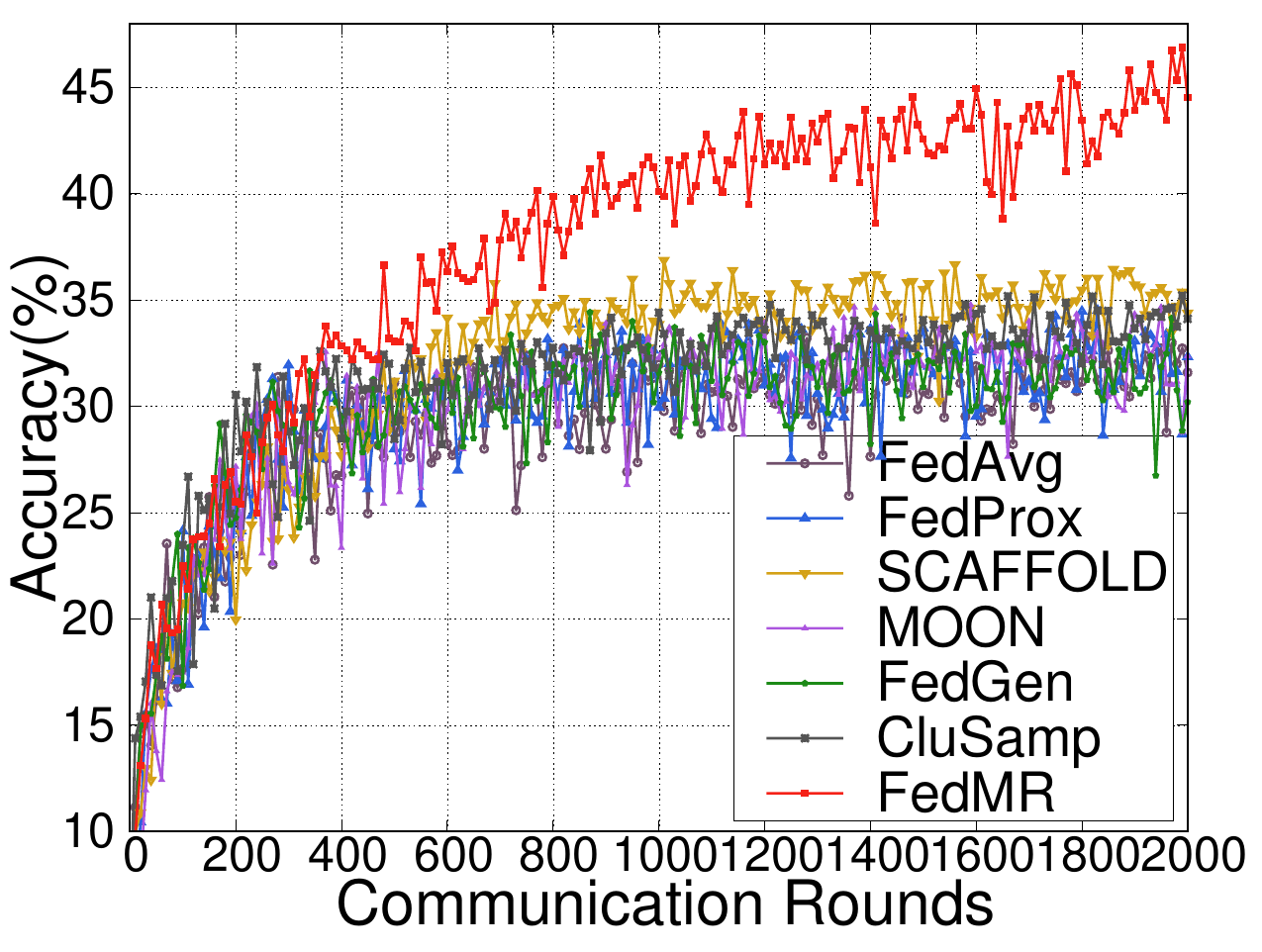}
		\label{fig:cifar-10-0.1}
	}\hspace{-0.15in}
	\subfigure[CIFAR-100  with $\alpha=0.5$]{
		\centering
		\includegraphics[width=0.25\textwidth]{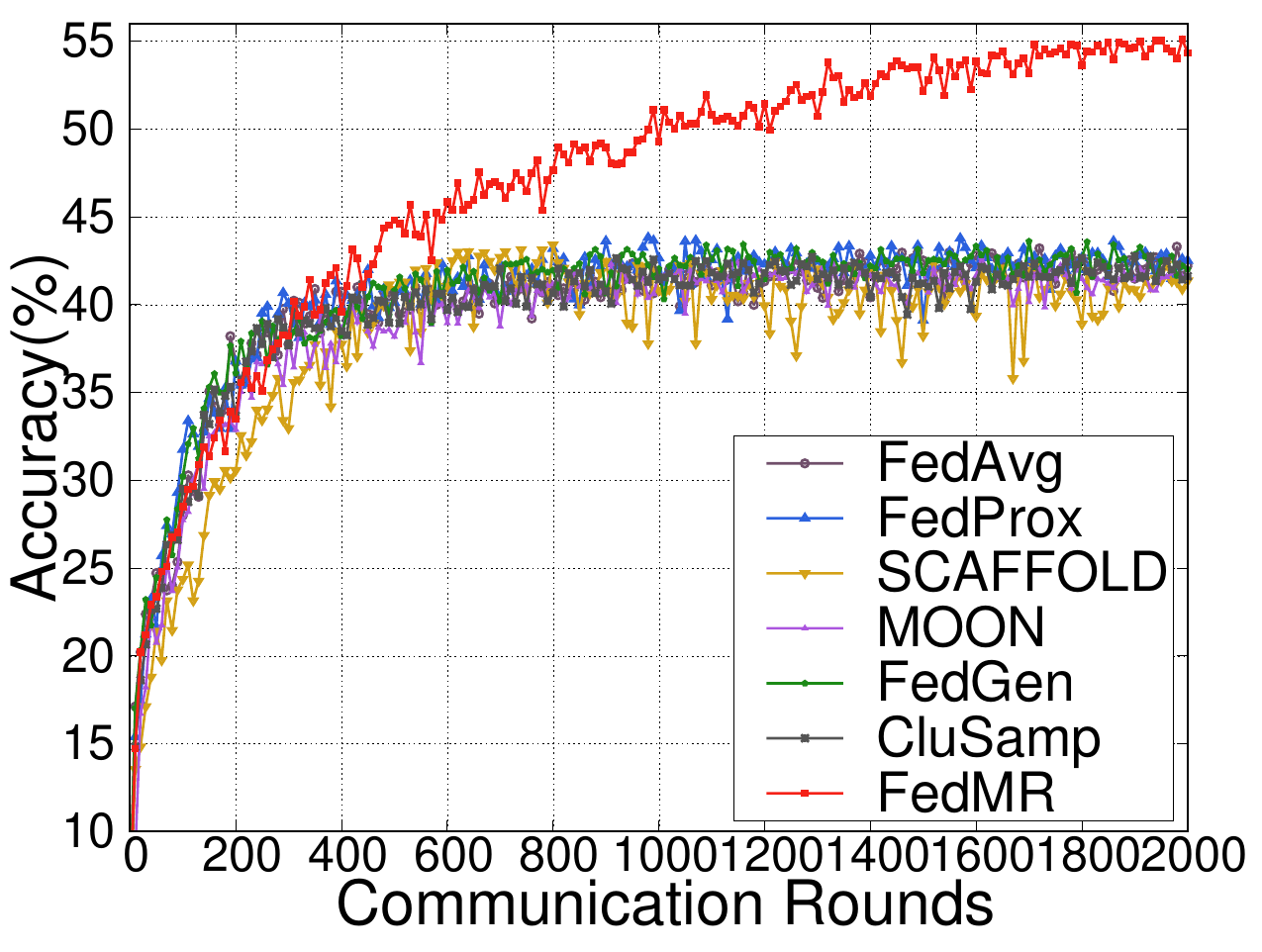}
		\label{fig:cifar-10-0.5}
	}\hspace{-0.15in}
	\subfigure[CIFAR-100 with $\alpha=1.0$]{
		\centering
		\includegraphics[width=0.25\textwidth]{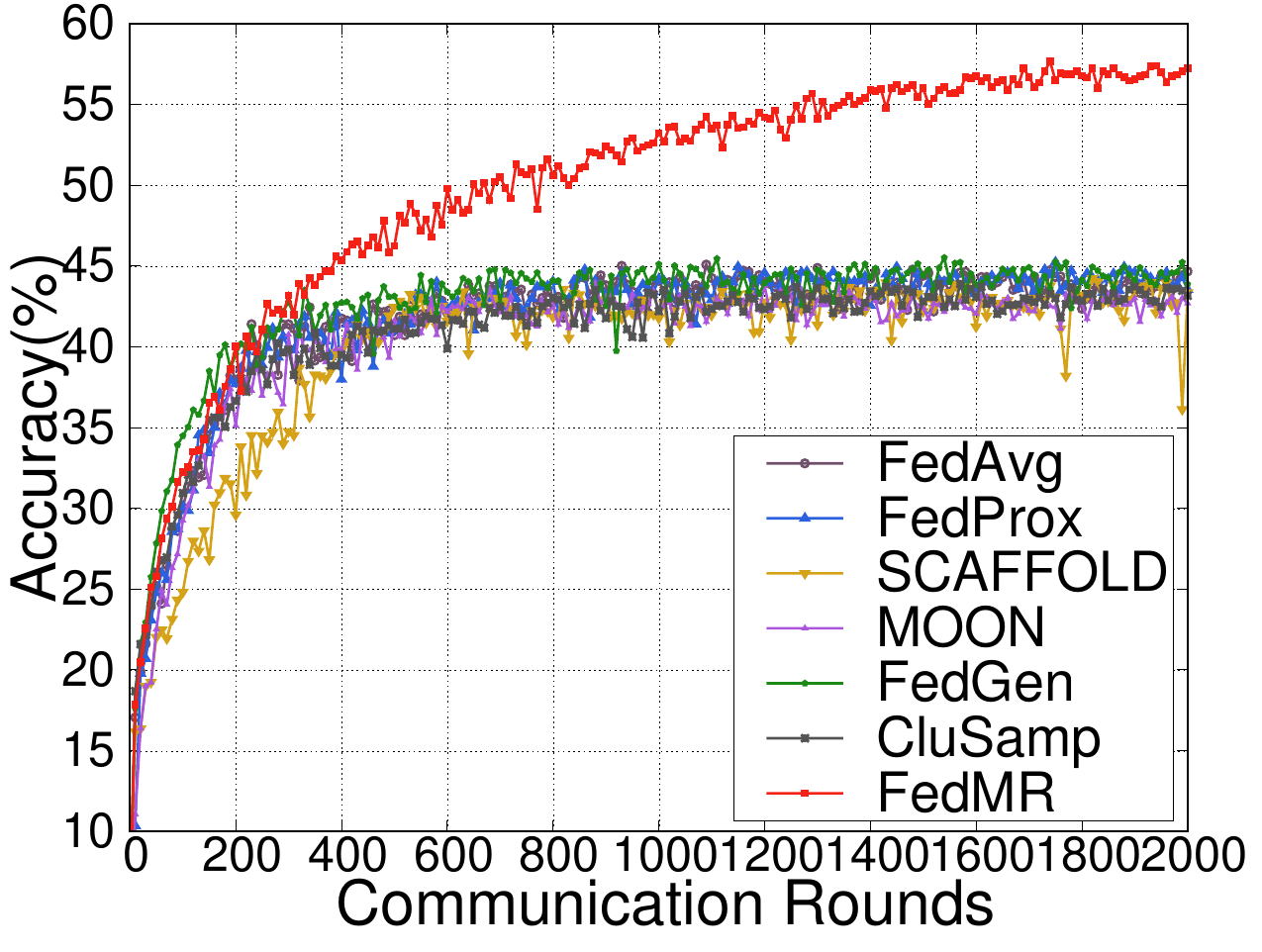}
		\label{fig:cifar-10-1.0}
	}\hspace{-0.15in}
	\subfigure[CIFAR-100 with IID]{
		\centering
		\includegraphics[width=0.25\textwidth]{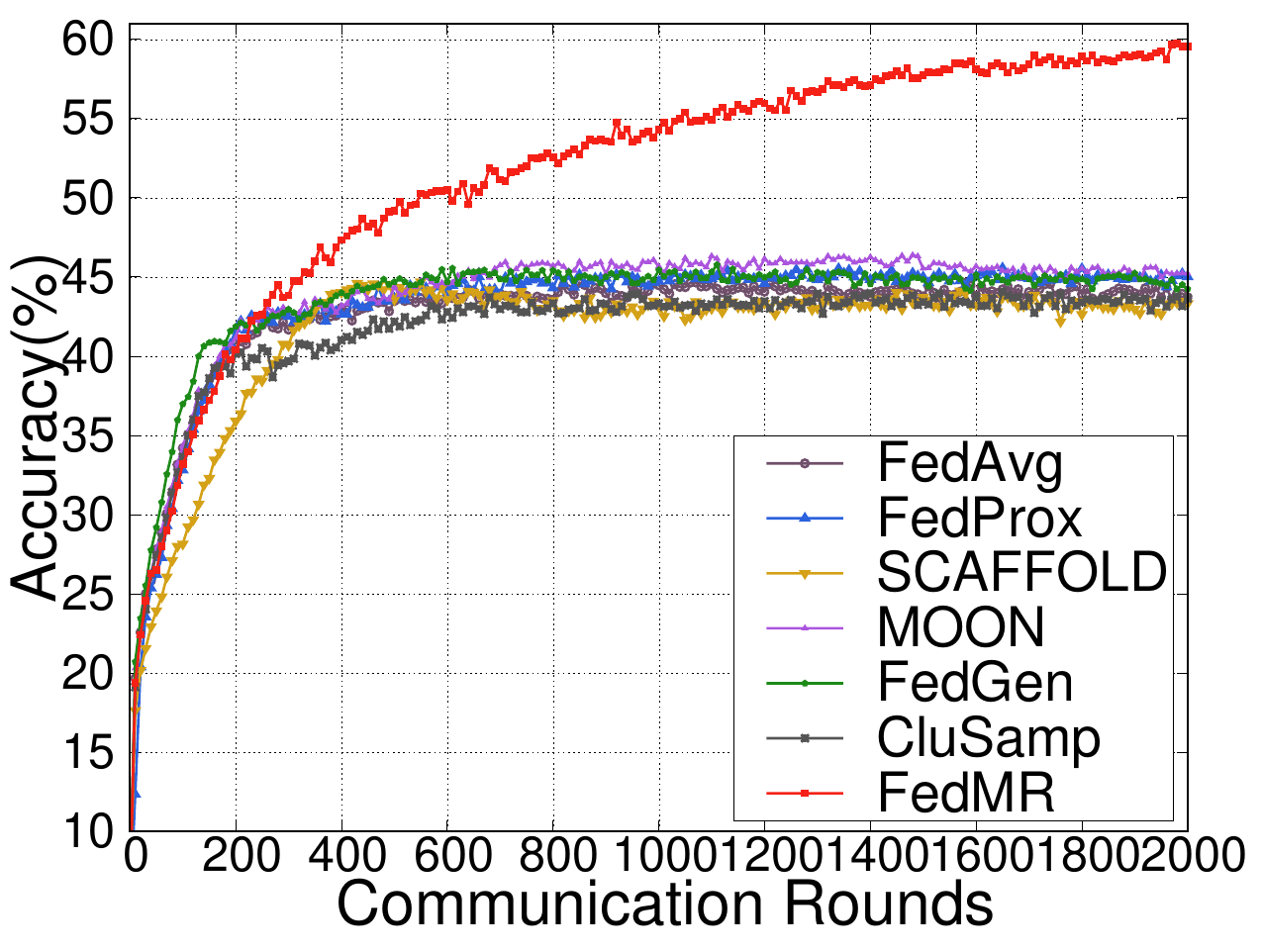}
		\label{fig:femnist}
	}
	\vspace{-0.1in}
	\caption{Learning curves of different FL methods based on the ResNet-20 model for CIFAR-100 dataset.}
	\label{fig:accuacy}
	\vspace{-0.15in}
\end{figure*}

\subsection{Performance Comparison}
We compared the performance of our FedMR approach with six SOTA baselines. 
For  datasets CIFAR-10 and CIFAR-100, we considered both IID and non-IID scenarios
(with $\alpha=0.1,0.5,1.0$, respectively).

\subsubsection{Comparison of Test Accuracy}
Table \ref{tab:acc} compares FedMR with the  SOTA FL methods considering both 
non-IID and IID scenarios based on three  different DL models. 
The first   two columns denote
the model type and dataset type, respectively. 
Note that to enable fair comparison, we cluster the test accuracy results 
generated by the FL methods based on the same type of local models. 
The  third column shows  different  distribution settings for 
client data, indicating the  data heterogeneity of  clients. 
The fourth column has seven sub-columns, which present the test accuracy 
information together with its standard deviation
for all the investigated FL methods, respectively. 

From Table \ref{tab:acc}, we can observe that FedMR can achieve the highest
test accuracy in all the scenarios regardless of model type, dataset type and data heterogeneity. For datasets CIFAR-10 and CIFAR-100, we can find that FedMR outperforms the six baseline
methods significantly in both non-IID and IID scenarios. 
For example, when dealing with a non-IID  CIFAR-10 scenario ($\alpha=0.1$)
using ResNet-20-based models, FedMR achieves  test  accuracy with 
an average of 62.09\%, while  the second highest average test accuracy obtained by 
SCAFFOLD is only 50.46\%.
Note that the performance of  FedMR on 
FEMNIST is not as notable as the one on both CIFAR-10 and CIFAR-100. 
This is mainly because the classification
task on FEMNIST is 
much 
simpler than the ones 
applied on datasets CIFAR-10 and CIFAR-100, which lead to the high test accuracy 
of the six baseline methods. However, even in this case, 
FedMR can still achieve the best test accuracy among all the  investigated FL methods. 

\subsubsection{Comparison of Model Convergence}

Figure~\ref{fig:accuacy} presents the convergence trends 
of the seven  ResNet-20-based FL methods (including FedMR) on the CIFAR-100 dataset.
Note that here the training of FedMR is based on our proposed 
two-stage training scheme, where the first stage uses 100 FL training rounds to achieve a pre-trained  model. 
Here, to enable  fair comparison, the test accuracy  of
 FedMR at some 
FL training round is  calculated by an intermediate global model, which is an aggregated version of all the local models  within that round. 
The four sub-figures show the results for different data distributions of
clients. 
From this figure, we can find that FedMR outperforms the other six FL methods consistently 
in both non-IID and IID scenarios. This is mainly because FedMR can easily escape from the stuck-at-local-search due to the model recombination operation in each FL round. Moreover,
due to the fine-grained local training, 
we can observer that the learning curves in each sub-figure is much smoother than the ones of  other FL methods. 
We also conducted the comparison for CNN- and VGG-16-based FL methods, and found the 
similar observations from  them.  Please refer to  Appendix for more details.

\subsubsection{Comparison of Communication Overhead}

Let $K$ be the number of clients that participate each FL training round.
Since the cloud server  needs to send $K$ recombined models to $K$ clients and receive 
$K$ trained local models from $K$ clients in an FL round, the communication overhead of FedMR is $2K$ models in each round, which is the same as FedAvg, FedPeox, and 
 CluSamp.
Since the cloud server needs to dispatch an extra global control variable to each client and clients also need to update and upload these global control variables to the cloud server, the communication overhead of SCAFFOLD is $2K$ models plus $2K$ global control variables in each FL training
round. 
Unlike the other FL methods, the cloud server of FedGen needs to dispatch 
an additional built-in generator to the selected clients, 
the communication overhead of FedGen in each FL training round is 
$2K$ models plus $K$ generators.
Base on the above 
analysis, we can find that FedMR does not 
cause  any extra  communication overhead. Therefore, FedMR requires 
the least  communication overhead among all the investigated FL methods in each FL training round.
Note that, as shown in Figure \ref{fig:accuacy}, although FedMR 
requires more rounds to achieve the highest test accuracy, to achieve the highest accuracy of other FL methods,  FedMR  generally requires less FL rounds. 
In other words, to achieve the same test accuracy, FedMR  requires much 
less overall 
communication overhead.

\subsection{Ablation Study}
To demonstrate the effectiveness of key components
of our FedMR approach, we conducted two ablation studies for model recombination and 
our two-stages FedMR training scheme, respectively.

\subsubsection{Model Recombination}
\begin{figure}[h]
	\begin{center}
		\includegraphics[width=2.2 in]{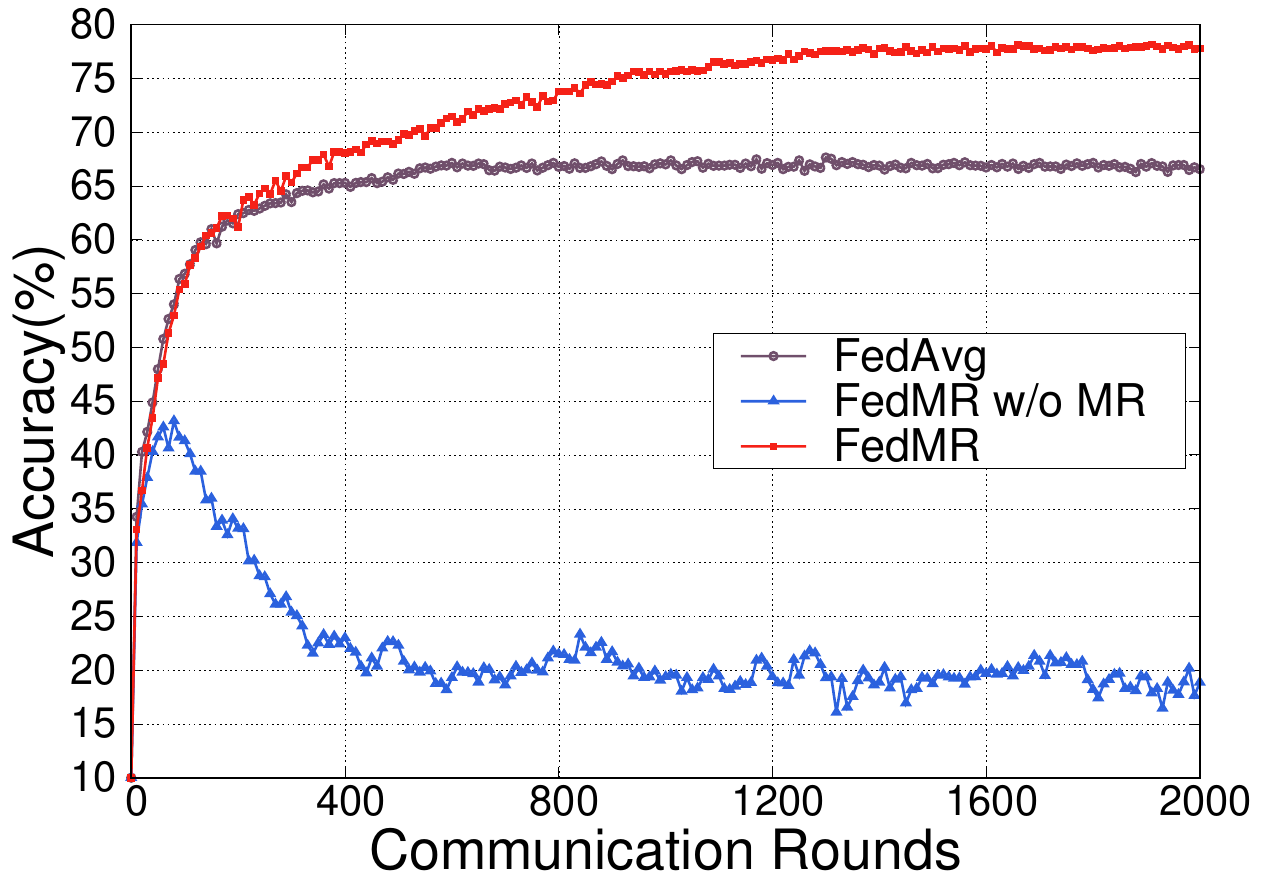}
		\caption{Ablation studies for model recombination.} \label{fig:abl_mr}
	\end{center}
	\vspace{-0.3in}
\end{figure}

To evaluate the effectiveness
of  model recombination, we developed  
a  variant of FedMR  named  {\bf ``FedMR w/o MR''}
without taking model recombination into account. 
In other words, in {\bf ``FedMR w/o MR''}
local models are directly dispatched to 
clients in a random manner with neither model aggregation nor model recombination. 
Figure~\ref{fig:abl_mr} presents the ablation study results on CIFAR-10 dataset
using ResNet-20-based FedMR, where the data on client are IID distributed.  
Note that all the FedMR results in Figure \ref{fig:abl_mr} are obtained from 
intermediate global models in differnt FL training rounds. 
From this figure, we can find that without using model recombination, the performance of
{\bf ``FedMR w/o MR''} degrades greatly.
In other words, the model recombination operation
is the key of  FedMR to achieve high classification performance. Moreover, since 
FedMR outperforms FedAvg significantly, it means that the model
recombination operation is more helpful than the model aggregation operation
in searching for better global models. 


\begin{figure}[h]
	\begin{center}
		\includegraphics[width=2.2in]{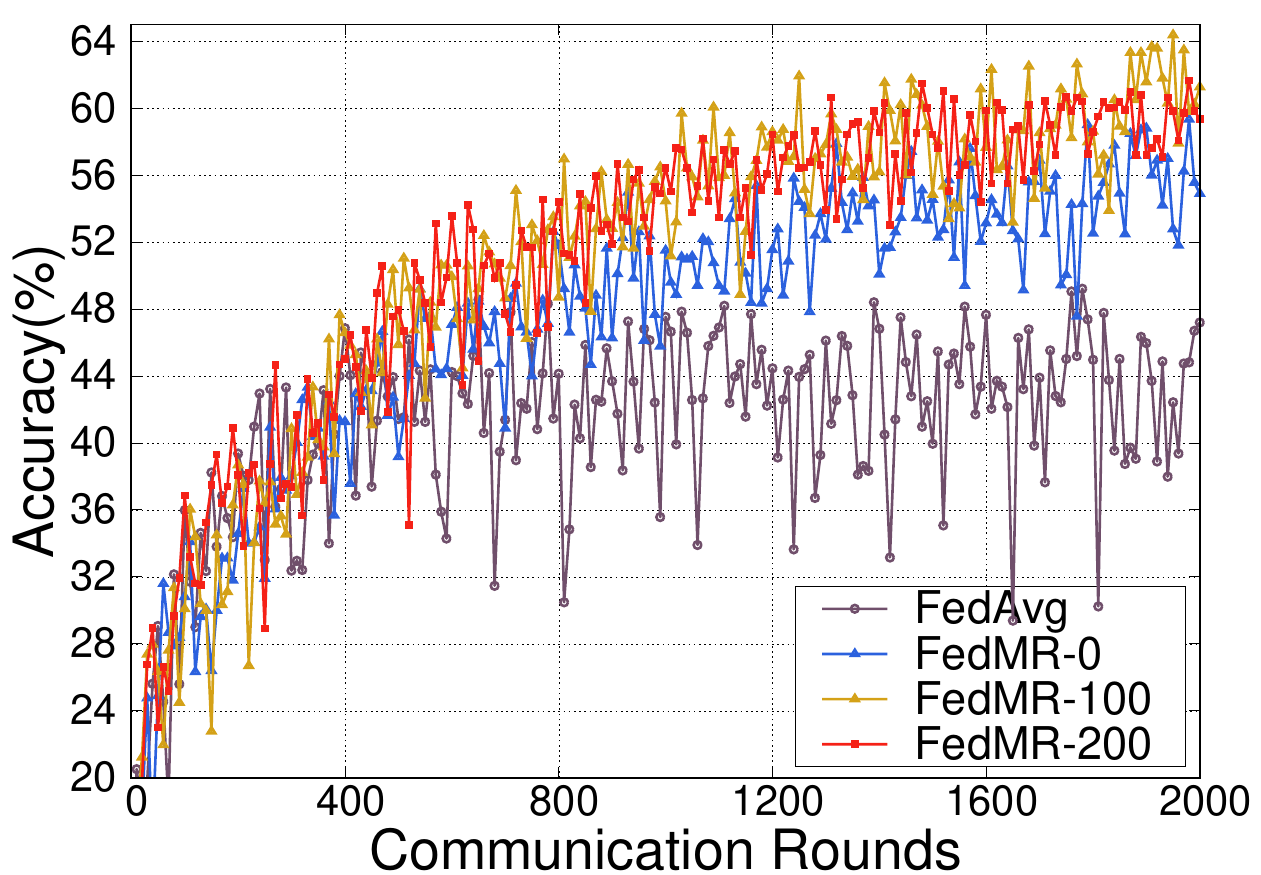}
		\caption{Ablation studies for the two-stage training scheme.} \label{fig:abl_twostage}
	\end{center}
\end{figure}

\subsubsection{Two-stage Training Scheme}
To demonstrate the effectiveness of our proposed two-stage training scheme, 
we conducted experiments on CIFAR-10 dataset using ResNet-20-based FedMR, 
where the data on client are non-IID  distributed ($\alpha=0.1$).  
Figure~\ref{fig:abl_twostage} presents the learning curves of FedAvg and FedMR, where 
FedMR adopts three 
 different two-stage training
 settings.
 Here,  we use the notation  ``FedMR-$n$''
 to denote that the first stage involves $n$ rounds of  model 
 aggregation-based local training to obtain a pre-trained global model, while 
 the remaining rounds conduct local training based on our proposed model recombination-based  method. 
From Figure~\ref{fig:abl_twostage}, we can observe that the two-stage training-based
FedMR methods (i.e., FedMR-100 and FedMR-200) achieve 
the best performance from the perspectives of test accuracy and convergence rate.

\subsection{Discussions}
\subsubsection{Privacy Preserving} 
Similar to traditional FedAvg-based FL methods, 
FedMR does not require clients to send their data to the cloud server, thus the 
data privacy can be mostly guaranteed by
the secure clients themselves. 
Someone may argue that  dispatching 
the recombined models to adversarial 
clients may  expose the privacy of some clients by 
attacking their specific layers. 
However, since our model recombination operation breaks 
the dependencies between model layers
and conducts the shuffling of layers among models, in practice
it is hard for adversaries to
restore the confidential data from 
a  fragmentary  recombined 
model without knowing the sources of layers. 
Note that, since the model update and dispatching operations of FedMR are the same as the ones of FedAvg-based  methods, any privacy-preserving techniques used in 
traditional FL method can  be easily integrated into FedMR to further reduce the privacy risks.

\subsubsection{Limitations}
As a novel FL paradigm, FedMR shows much better inference
performance than most SOTA FL methods. Although this paper proposed an efficient two-stage training scheme to accelerate the overall FL training processes, there
still exists numerous
chances (e.g., client selection strategies,  dynamic combination of model aggregation and model recombination operations)
to enable
further optimization on the current version of FedMR. 
Meanwhile, the current version of FedMR does not take the personalization into account, which is also a very important topic that is worthy of studying in the future.

\section{Conclusion}

Due to coarse-grained aggregation of FedAvg, when dealing with 
uneven data distribution among clients,
existing Federated Learning (FL) methods greatly suffer from 
the problem of low inference performance.
To address this problem, this paper presented a new FL paradigm named
FedMR, which enables different layers of local models to be trained on 
different clients in different FL  training 
rounds based on our proposed federated model recombination method. 
Since our approach allows both fine-grained model recombination 
and different  models for local training
in each FL training round, 
FedMR can  search for globally optimal models for all the clients in a more efficient
manner. Comprehensive experimental results show both
the effectiveness and pervasiveness
of our propose methods in term of inference accuracy and  convergence rate. 


\bibliographystyle{unsrt}

\appendix

\section{Supplementary Experimental Results}

\subsubsection{Experimental Results for Accuracy Comparison}
\begin{figure}[h]
	\centering
	\subfigure[CIFAR-10 with $\alpha=0.1$]{
		\centering
		\includegraphics[width=0.2\textwidth]{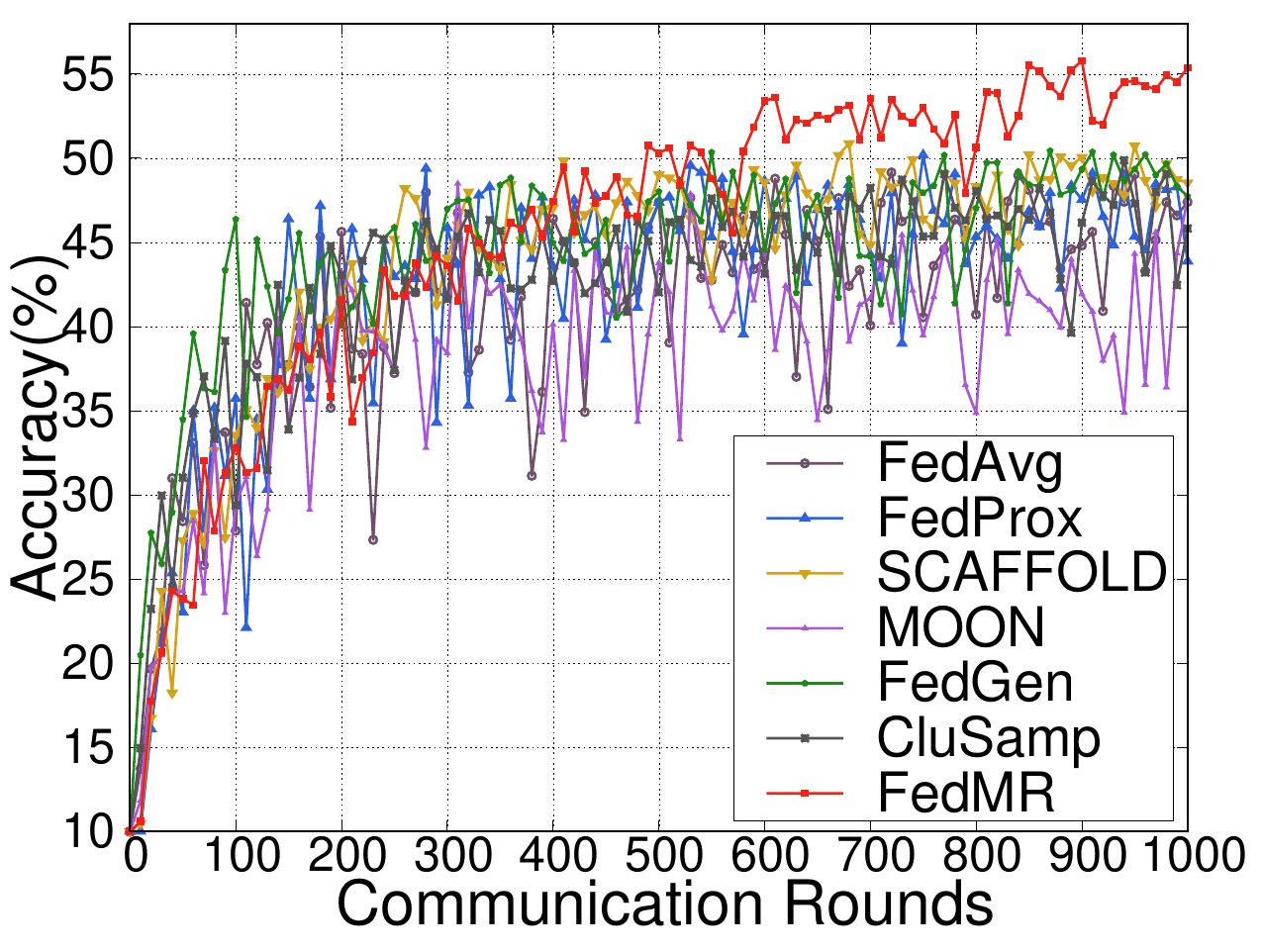}
	}\hspace{-0.0in}
	\subfigure[CIFAR-10  with $\alpha=0.5$]{
		\centering
		\includegraphics[width=0.2\textwidth]{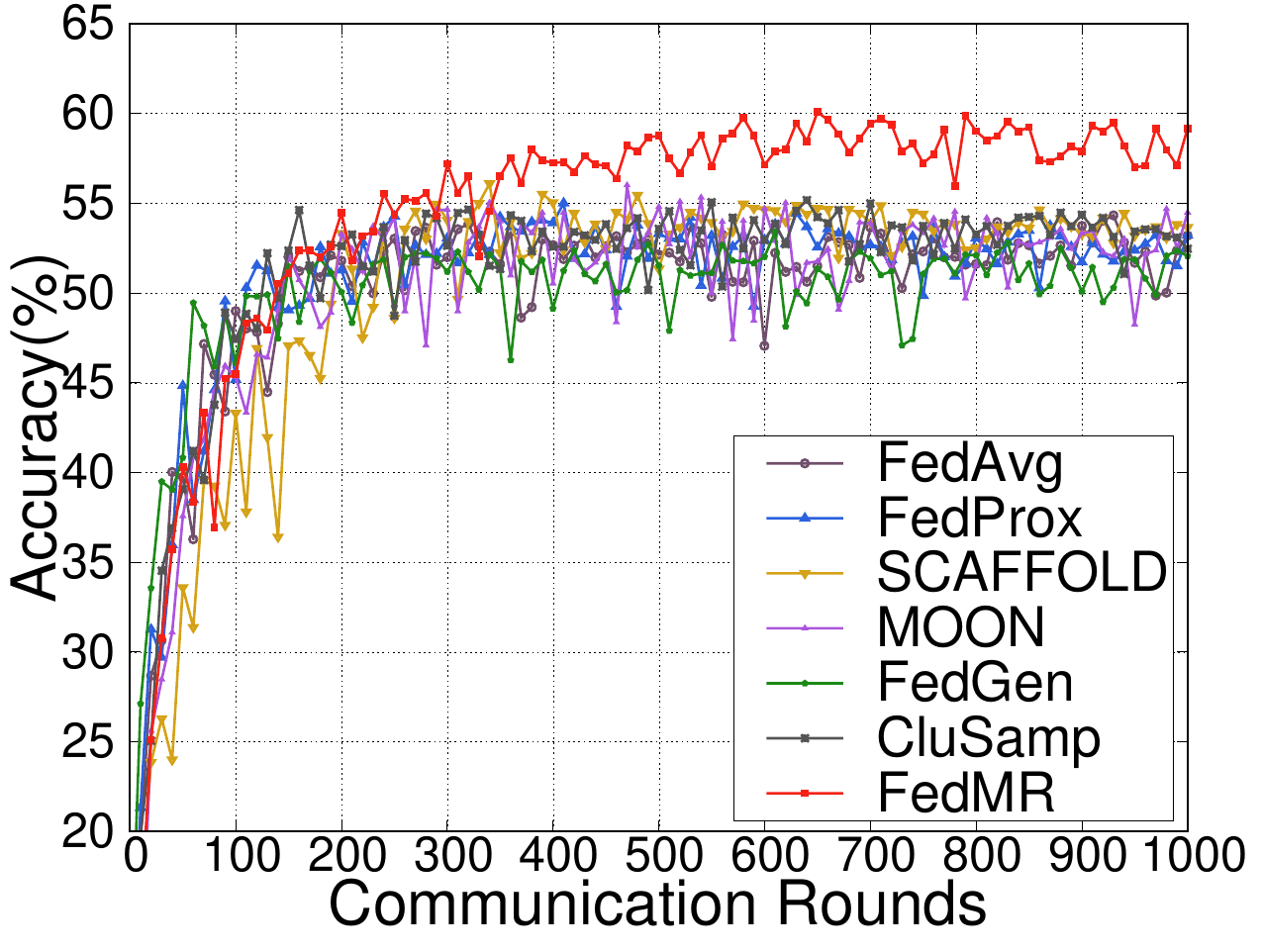}
	}\hspace{-0.0in}
	\subfigure[CIFAR-10 with $\alpha=1.0$]{
		\centering
		\includegraphics[width=0.2\textwidth]{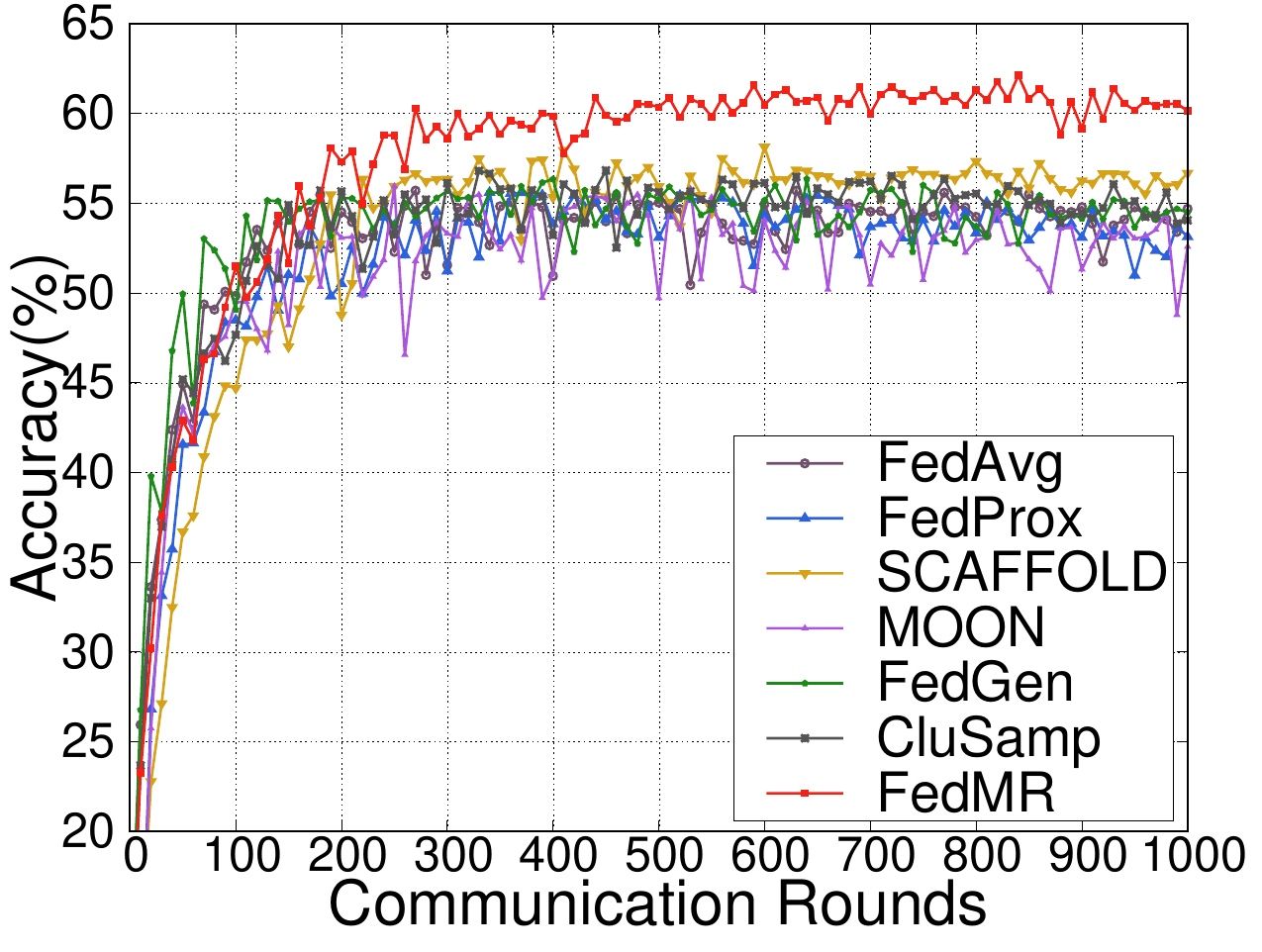}
	}\hspace{-0.0in}
	\subfigure[CIFAR-10 with IID]{
		\centering
		\includegraphics[width=0.2\textwidth]{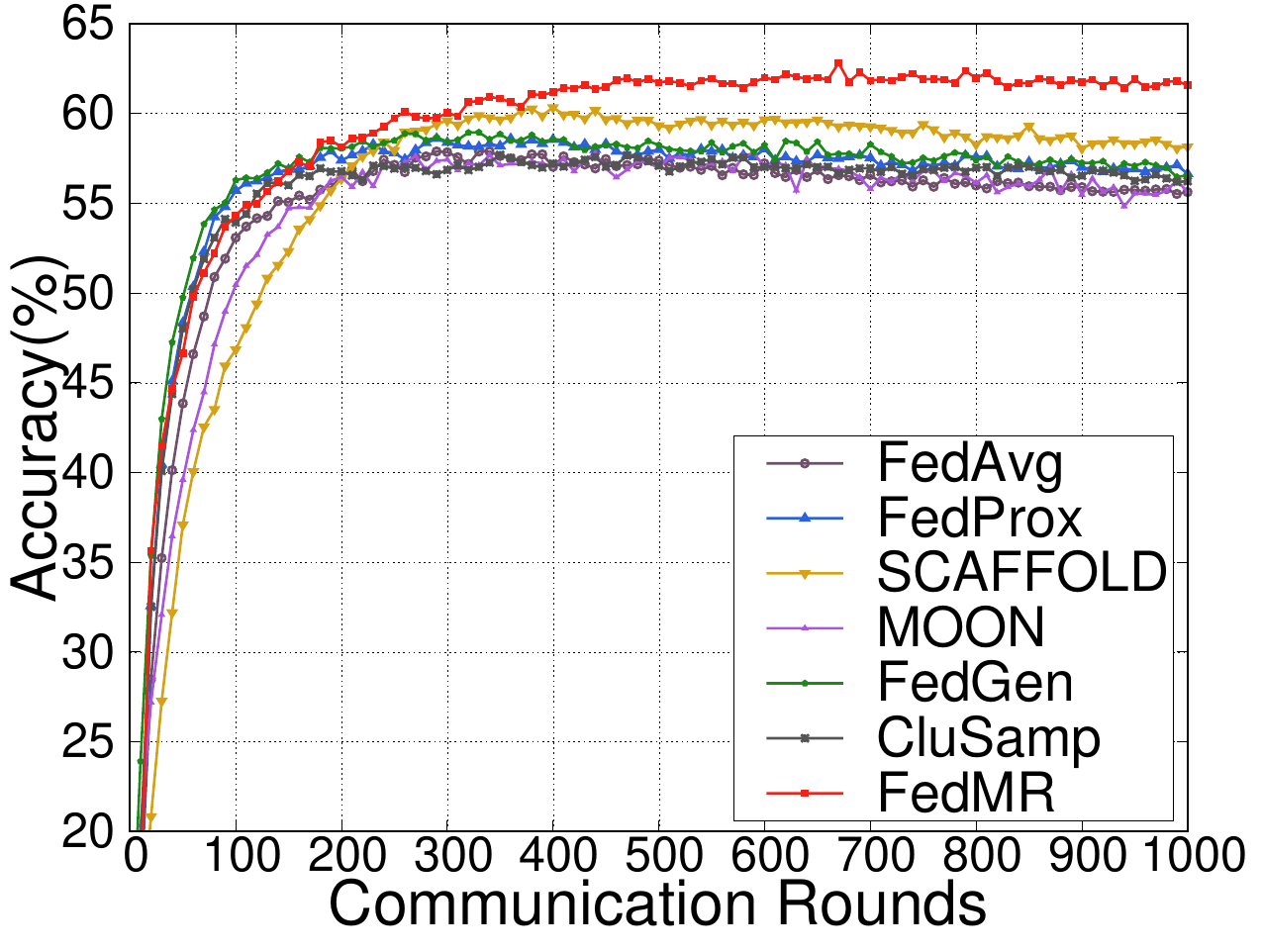}
	}
	\subfigure[CIFAR-100 with $\alpha=0.1$]{
		\centering
		\includegraphics[width=0.2\textwidth]{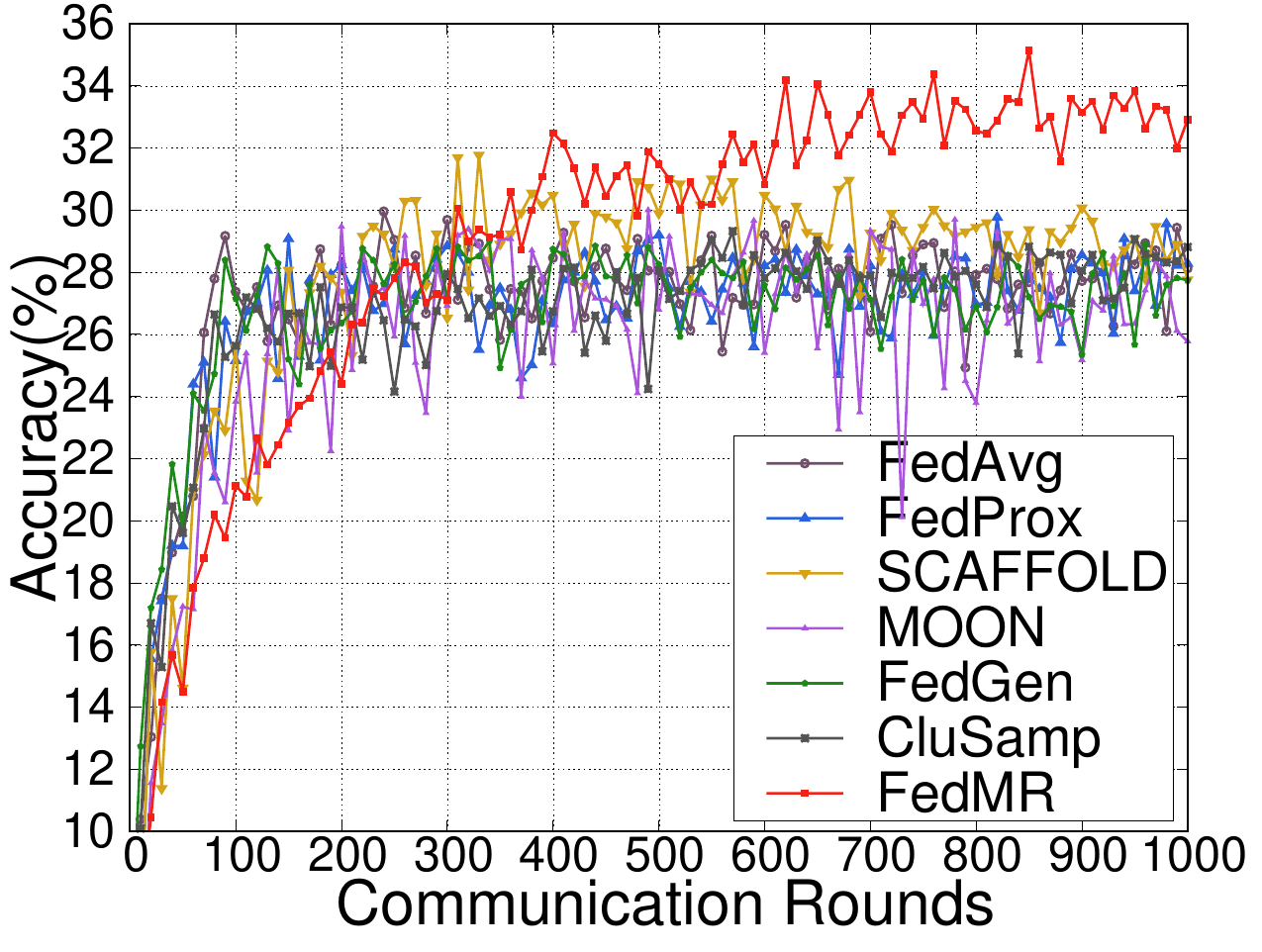}
	}\hspace{-0.0in}
	\subfigure[CIFAR-100  with $\alpha=0.5$]{
		\centering
		\includegraphics[width=0.2\textwidth]{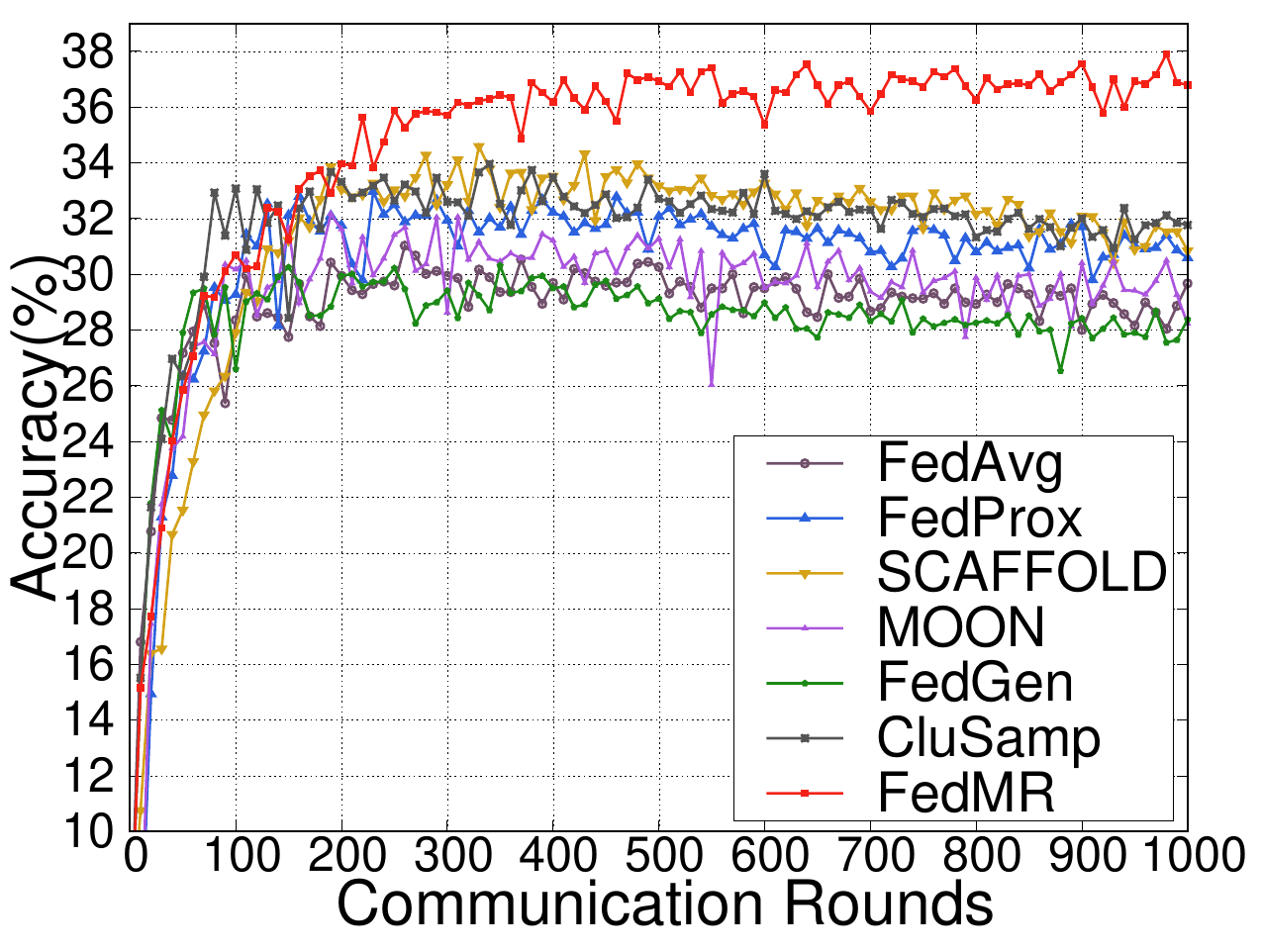}
	}\hspace{-0.0in}
	\subfigure[CIFAR-100 with $\alpha=1.0$]{
		\centering
		\includegraphics[width=0.2\textwidth]{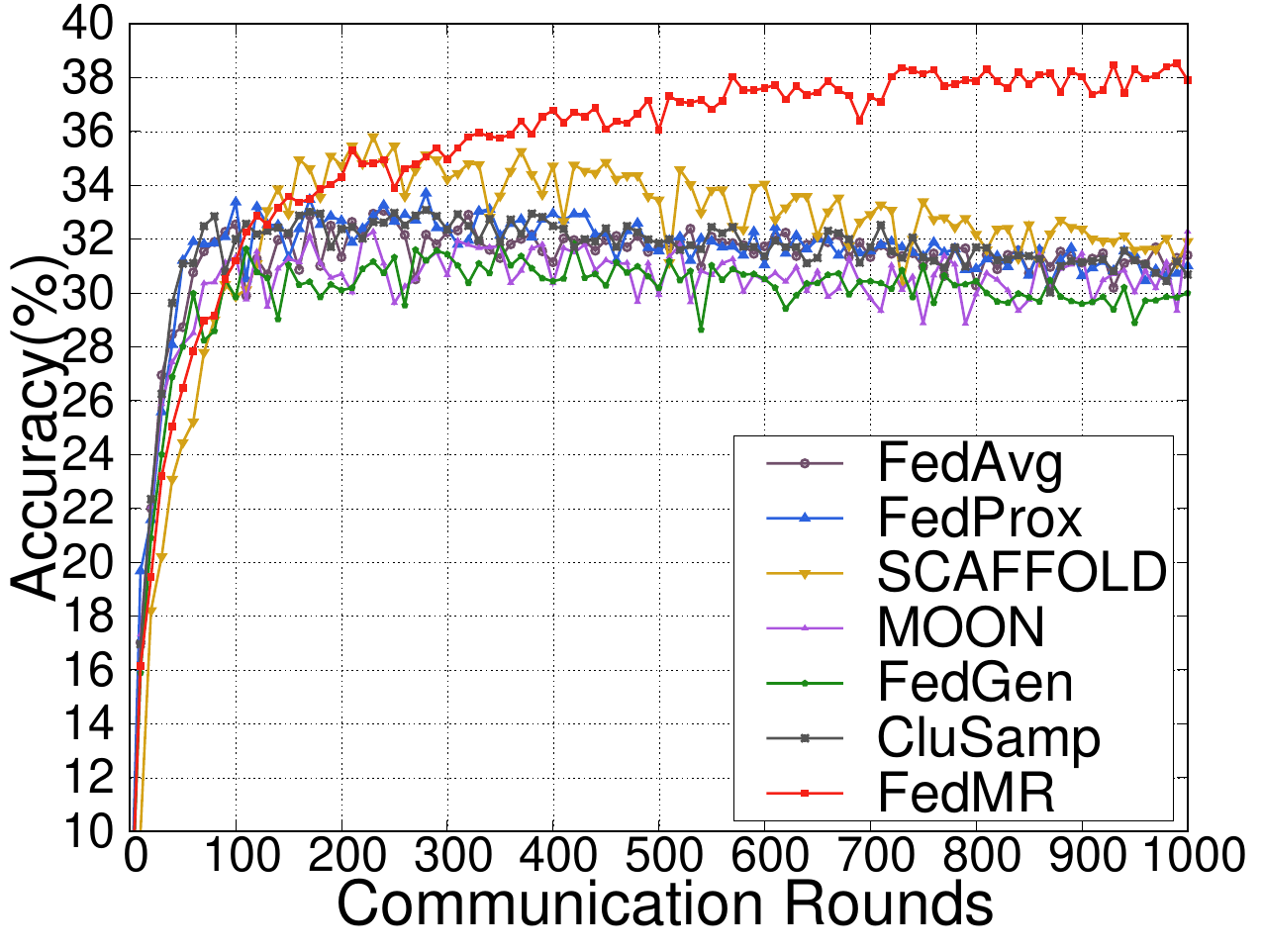}
	}\hspace{-0.0in}
	\subfigure[CIFAR-100 with IID]{
		\centering
		\includegraphics[width=0.2\textwidth]{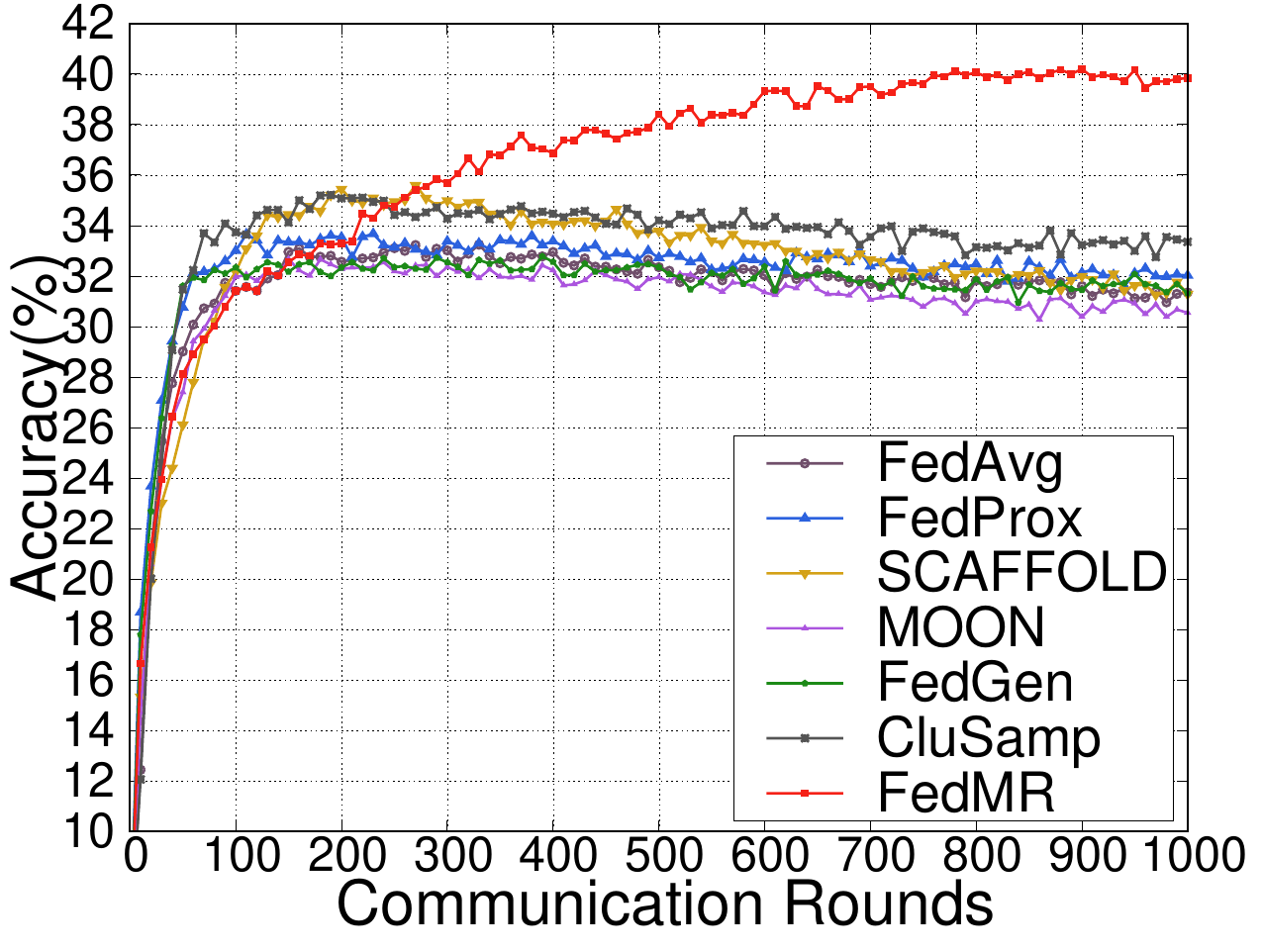}
	}
	\subfigure[FEMNIST]{
		\centering
		\includegraphics[width=0.2\textwidth]{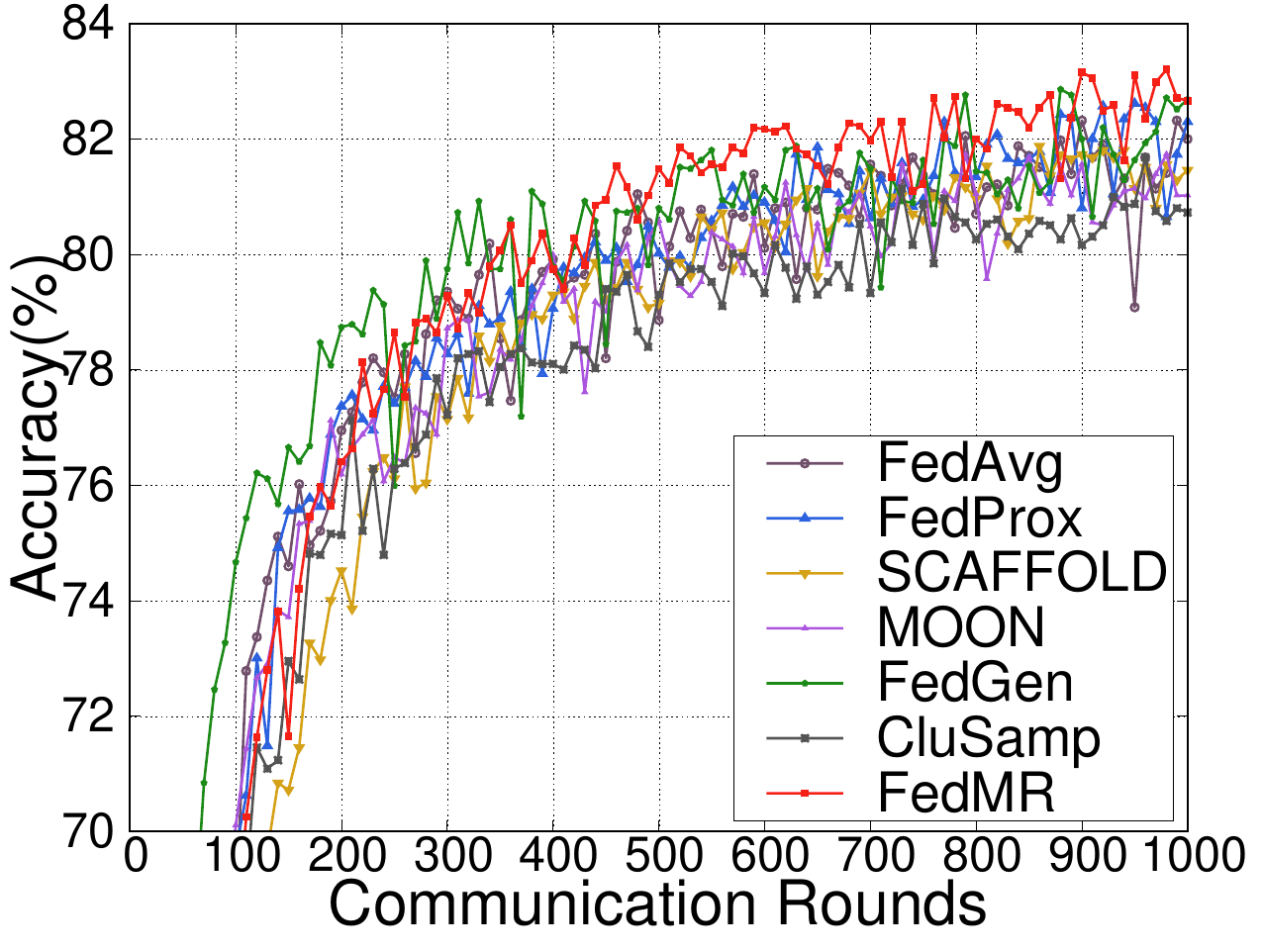}
	}
	\vspace{-0.0in}
	\caption{Learning curves of different FL methods based on the CNN model.}
	\label{fig:accuacy_cnn}
\end{figure}

\begin{figure}[b]
	\centering
	\subfigure[CIFAR-10 with $\alpha=0.1$]{
		\centering
		\includegraphics[width=0.2\textwidth]{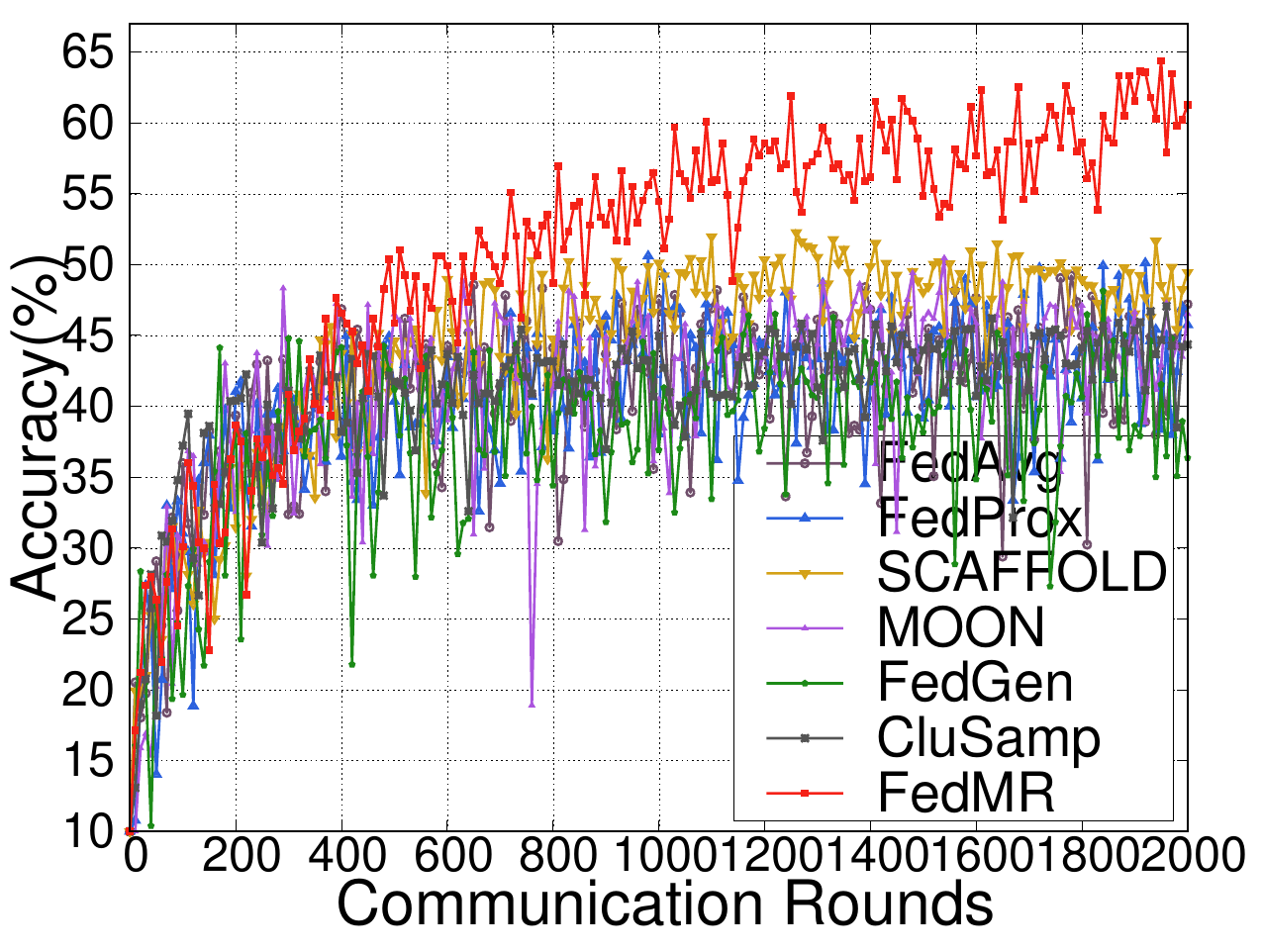}
	}
	\subfigure[CIFAR-10  with $\alpha=0.5$]{
		\centering
		\includegraphics[width=0.2\textwidth]{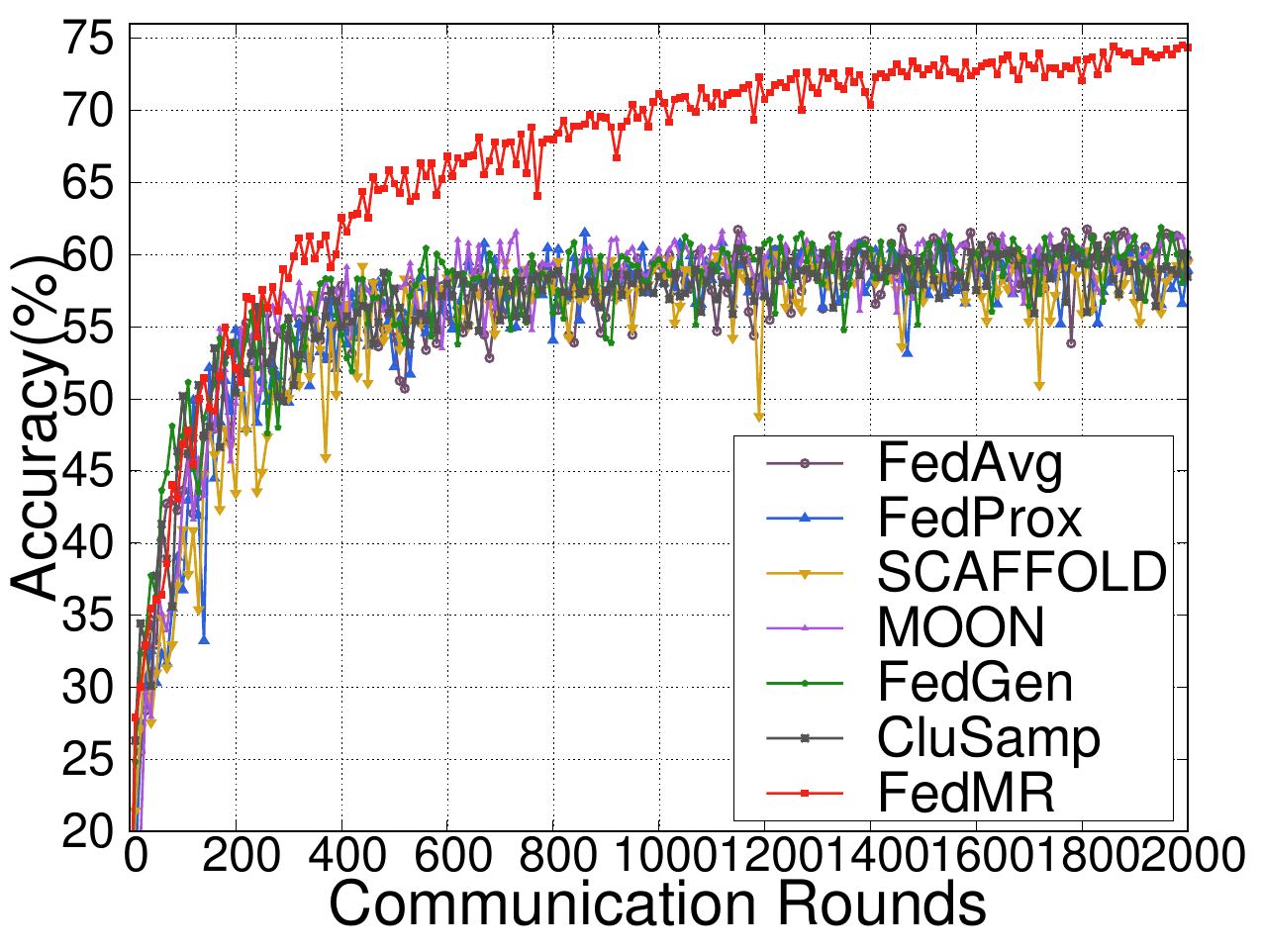}
	}
	\subfigure[CIFAR-10 with $\alpha=1.0$]{
		\centering
		\includegraphics[width=0.2\textwidth]{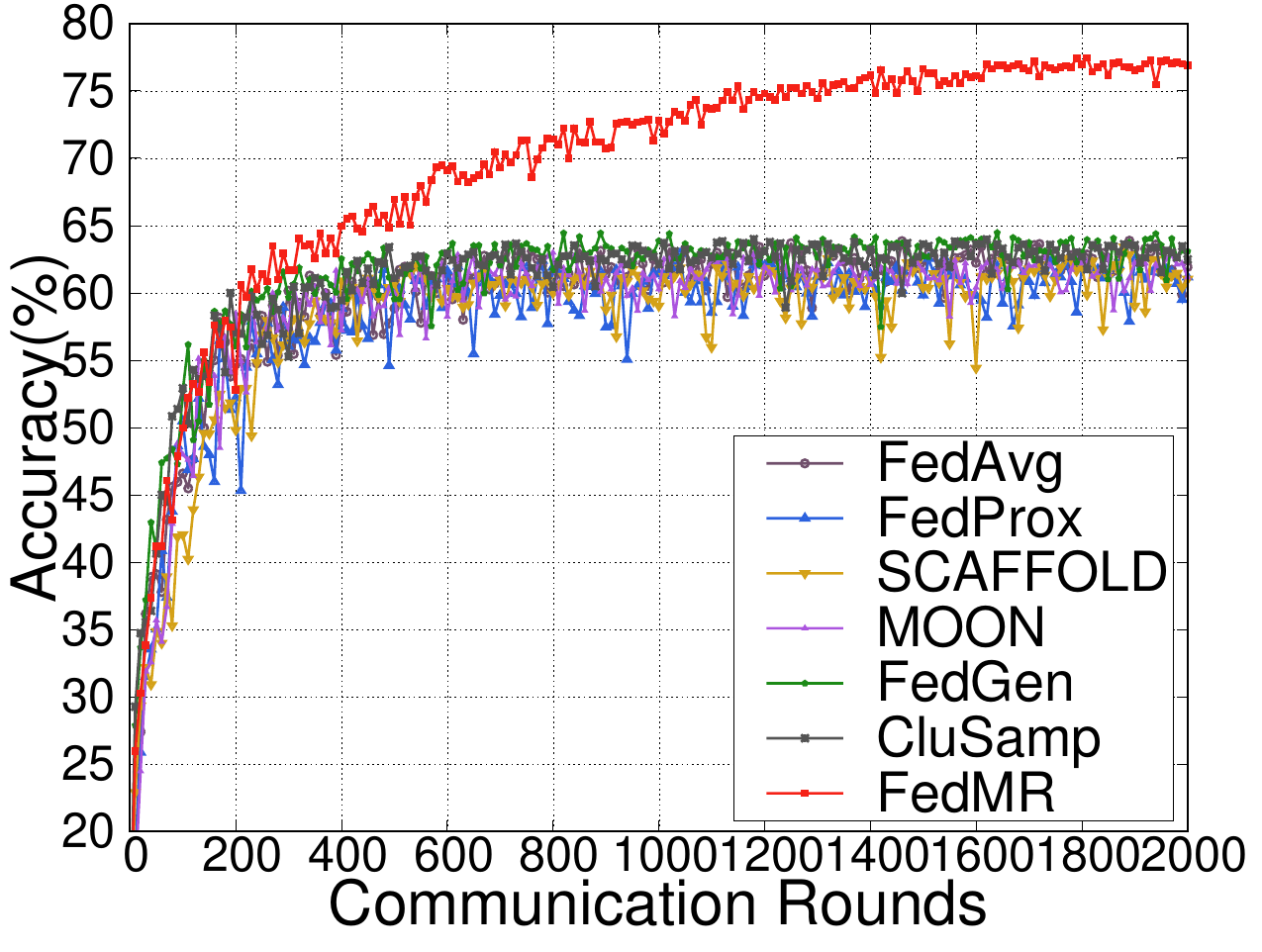}
	}
	\subfigure[CIFAR-10 with IID]{
		\centering
		\includegraphics[width=0.2\textwidth]{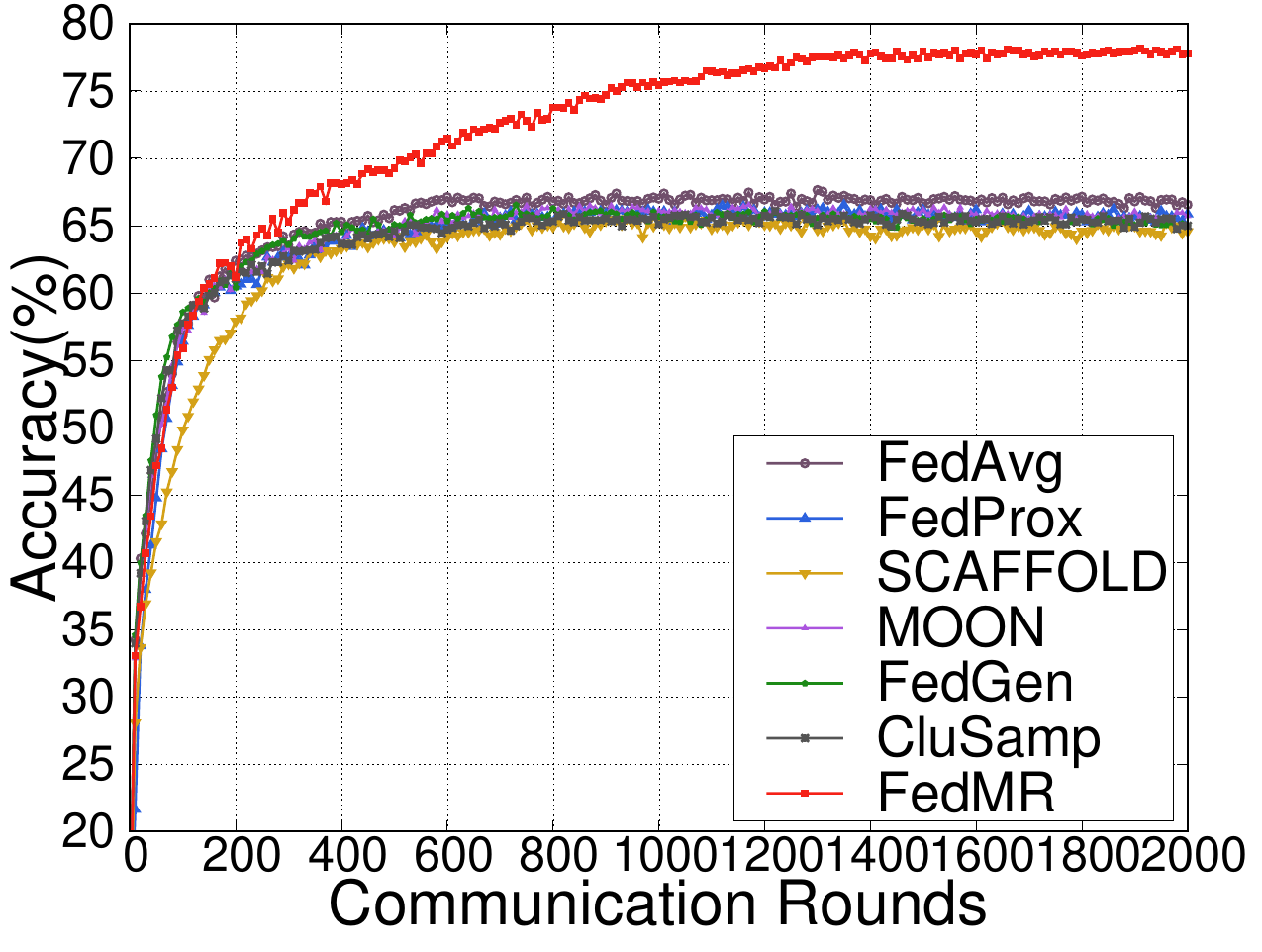}
	}
	\subfigure[CIFAR-100 with $\alpha=0.1$]{
		\centering
		\includegraphics[width=0.2\textwidth]{./fig/experiment/accuracy_resnet_cifar100_d0.1.pdf}
	}
	\subfigure[CIFAR-100  with $\alpha=0.5$]{
		\centering
		\includegraphics[width=0.2\textwidth]{./fig/experiment/accuracy_resnet_cifar100_d0.5.pdf}
	}
	\subfigure[CIFAR-100 with $\alpha=1.0$]{
		\centering
		\includegraphics[width=0.2\textwidth]{./fig/experiment/accuracy_resnet_cifar100_d1.0.pdf}
	}
	\subfigure[CIFAR-100 with IID]{
		\centering
		\includegraphics[width=0.2\textwidth]{./fig/experiment/accuracy_resnet_cifar100_iid.pdf}
	}
	\subfigure[FEMNIST]{
		\centering
		\includegraphics[width=0.2\textwidth]{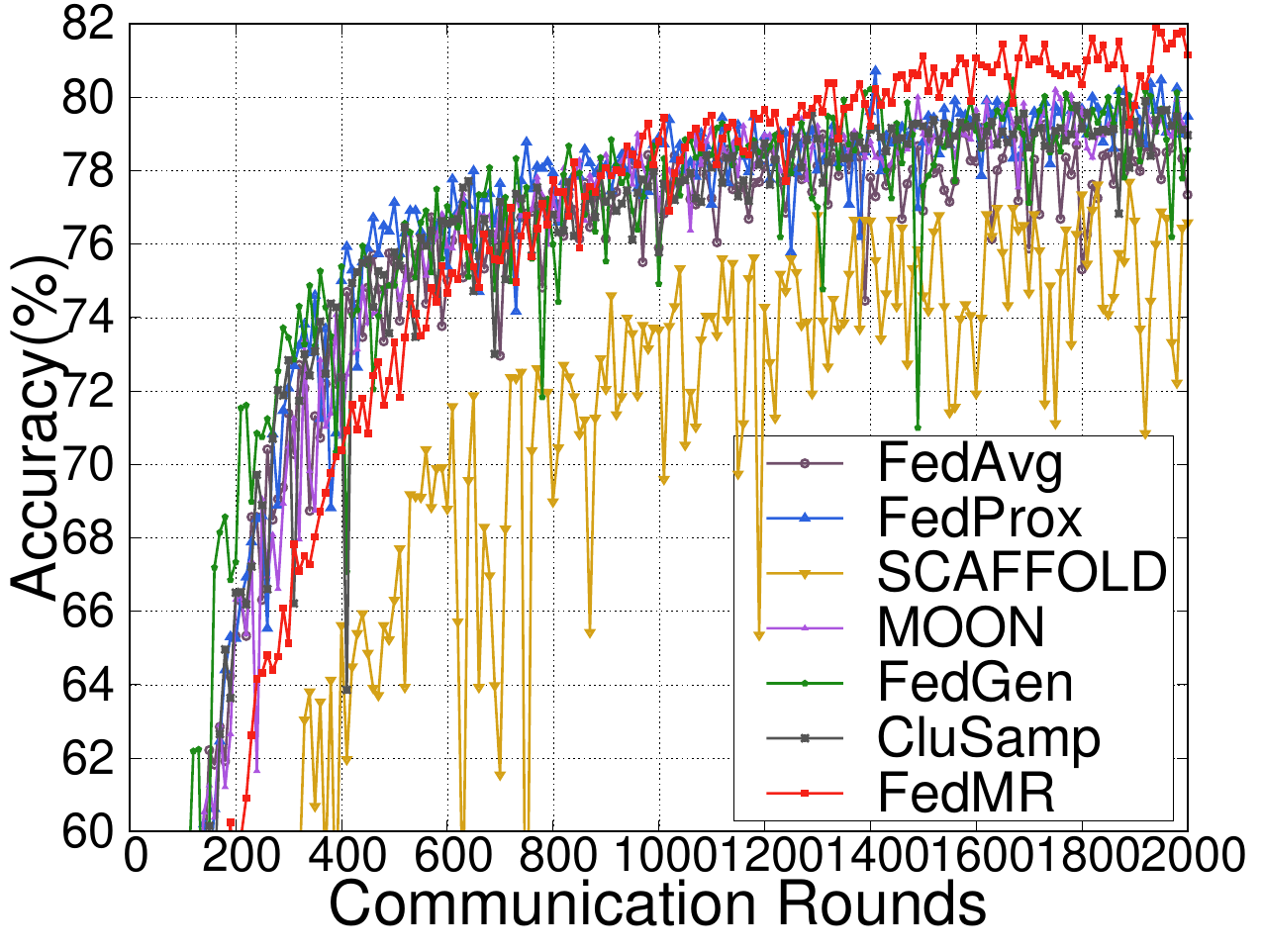}
	}
	\caption{Learning curves of different FL methods based on the ResNet-20 model.}
	\label{fig:accuacy_resnet}
\end{figure}

\begin{figure}[b]
	\centering
	\subfigure[CIFAR-10 with $\alpha=0.1$]{
		\centering
		\includegraphics[width=0.2\textwidth]{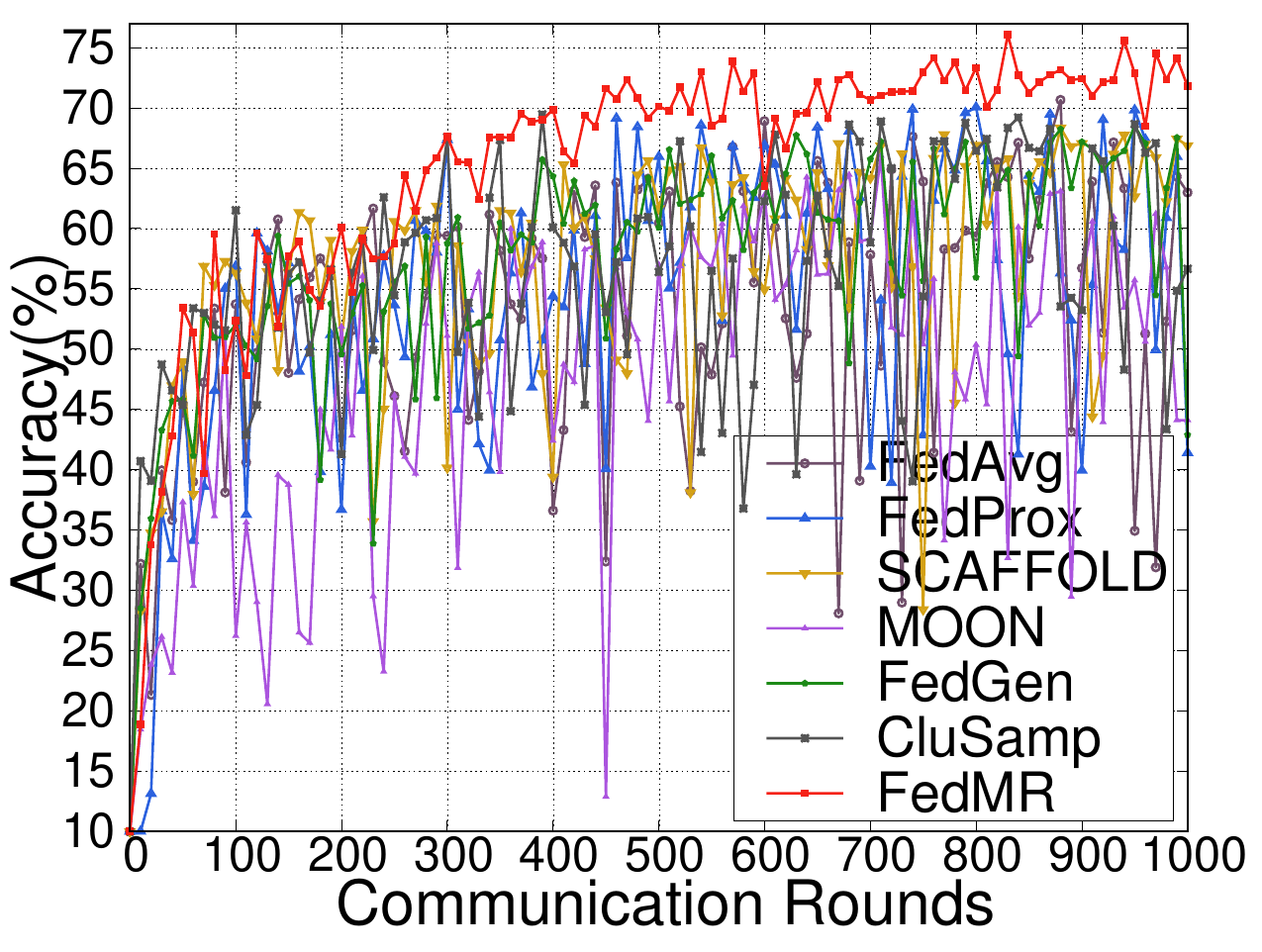}
	}
	\subfigure[CIFAR-10  with $\alpha=0.5$]{
		\centering
		\includegraphics[width=0.2\textwidth]{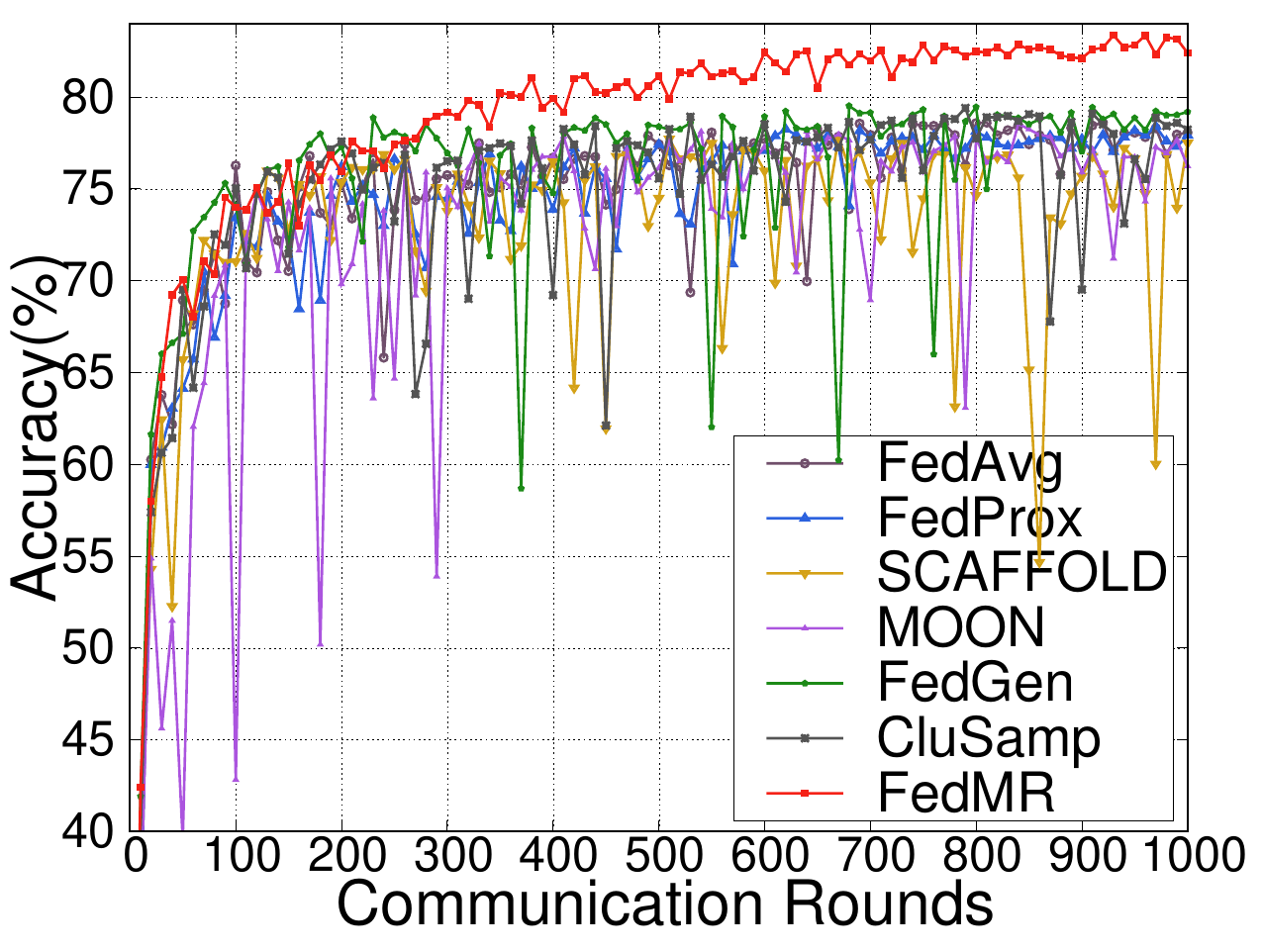}
	}
	\subfigure[CIFAR-10 with $\alpha=1.0$]{
		\centering
		\includegraphics[width=0.2\textwidth]{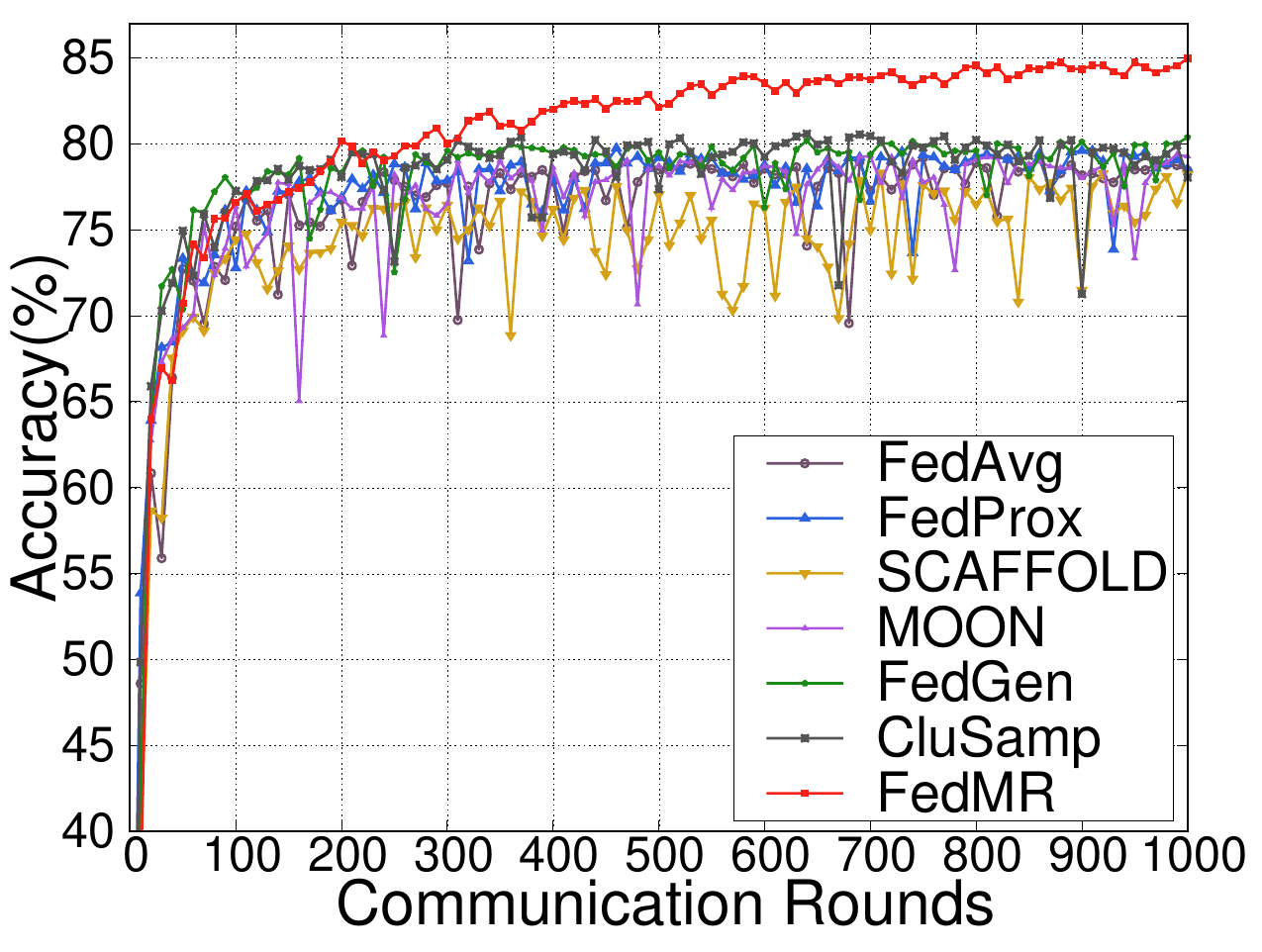}
	}
	\subfigure[CIFAR-10 with IID]{
		\centering
		\includegraphics[width=0.2\textwidth]{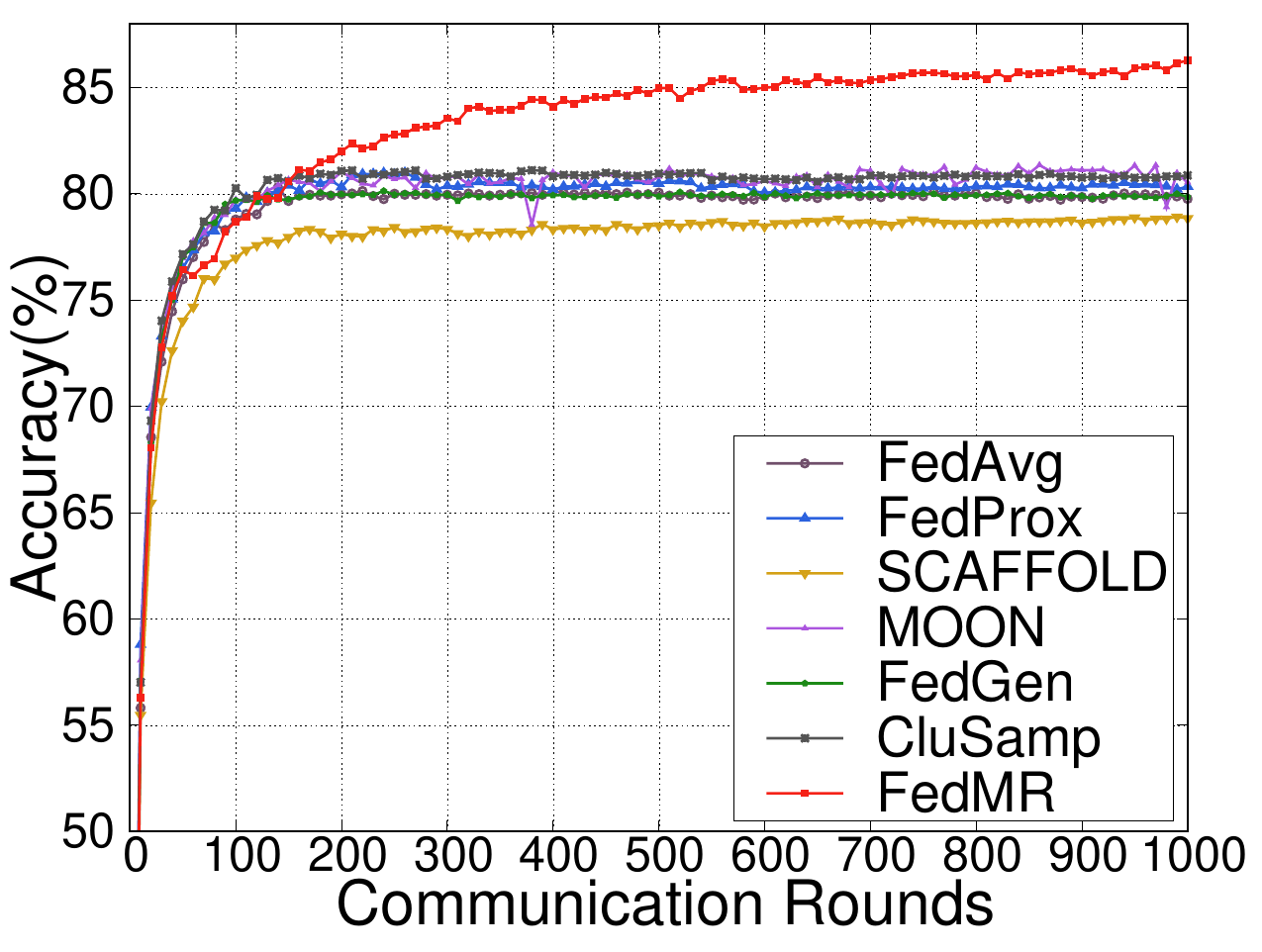}
	}
	\subfigure[CIFAR-100 with $\alpha=0.1$]{
		\centering
		\includegraphics[width=0.2\textwidth]{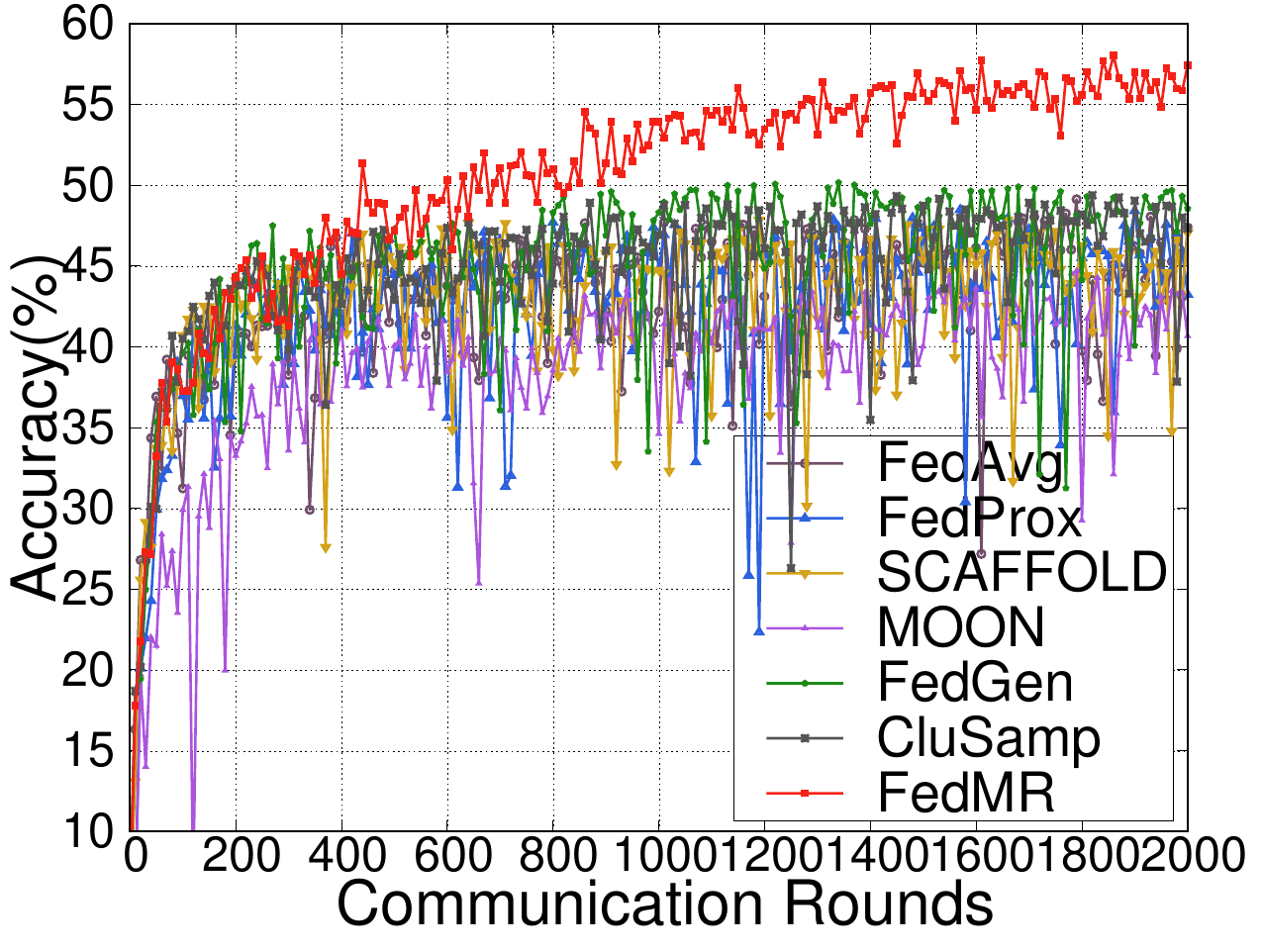}
	}
	\subfigure[CIFAR-100  with $\alpha=0.5$]{
		\centering
		\includegraphics[width=0.2\textwidth]{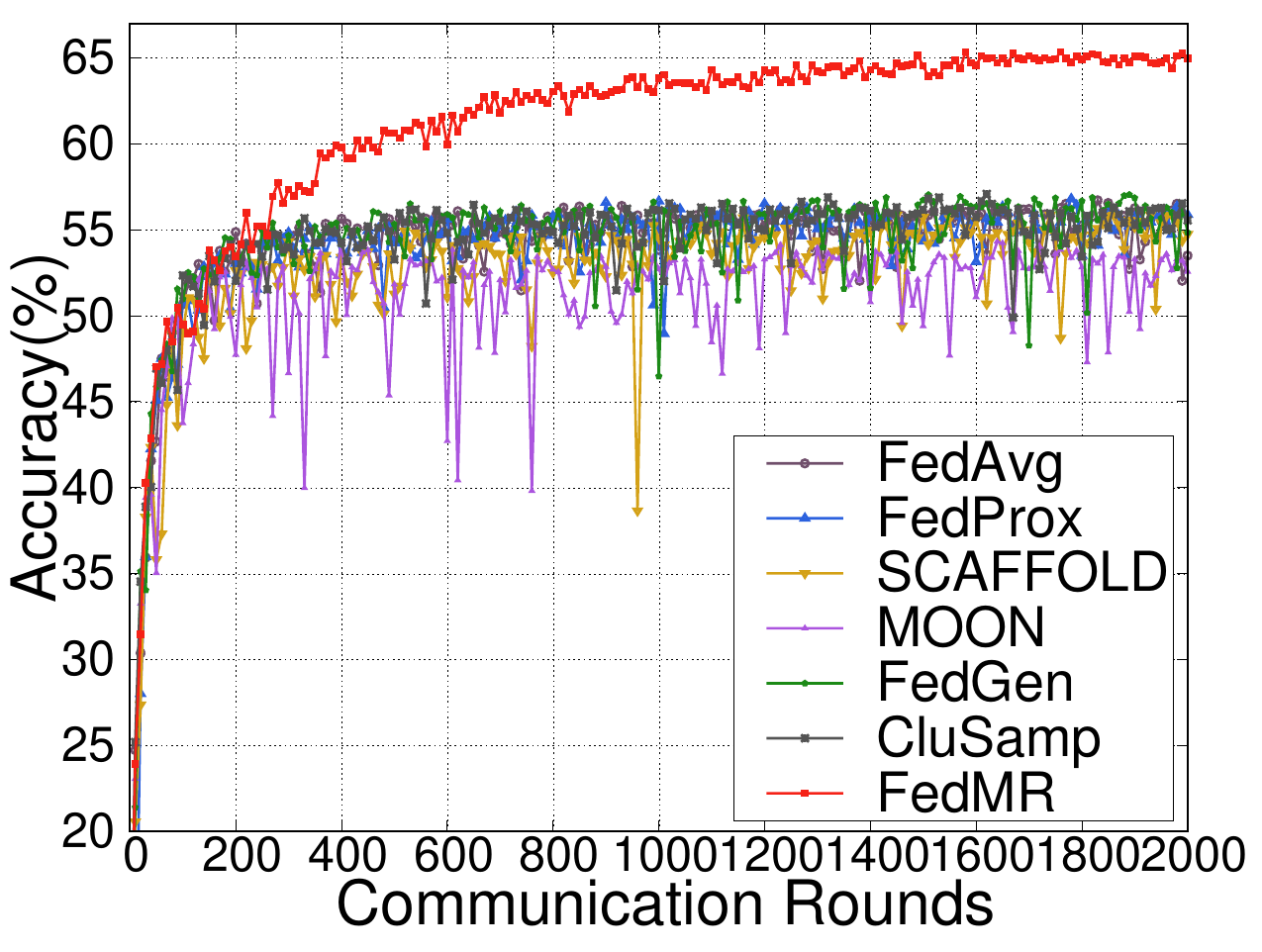}
	}
	\subfigure[CIFAR-100 with $\alpha=1.0$]{
		\centering
		\includegraphics[width=0.2\textwidth]{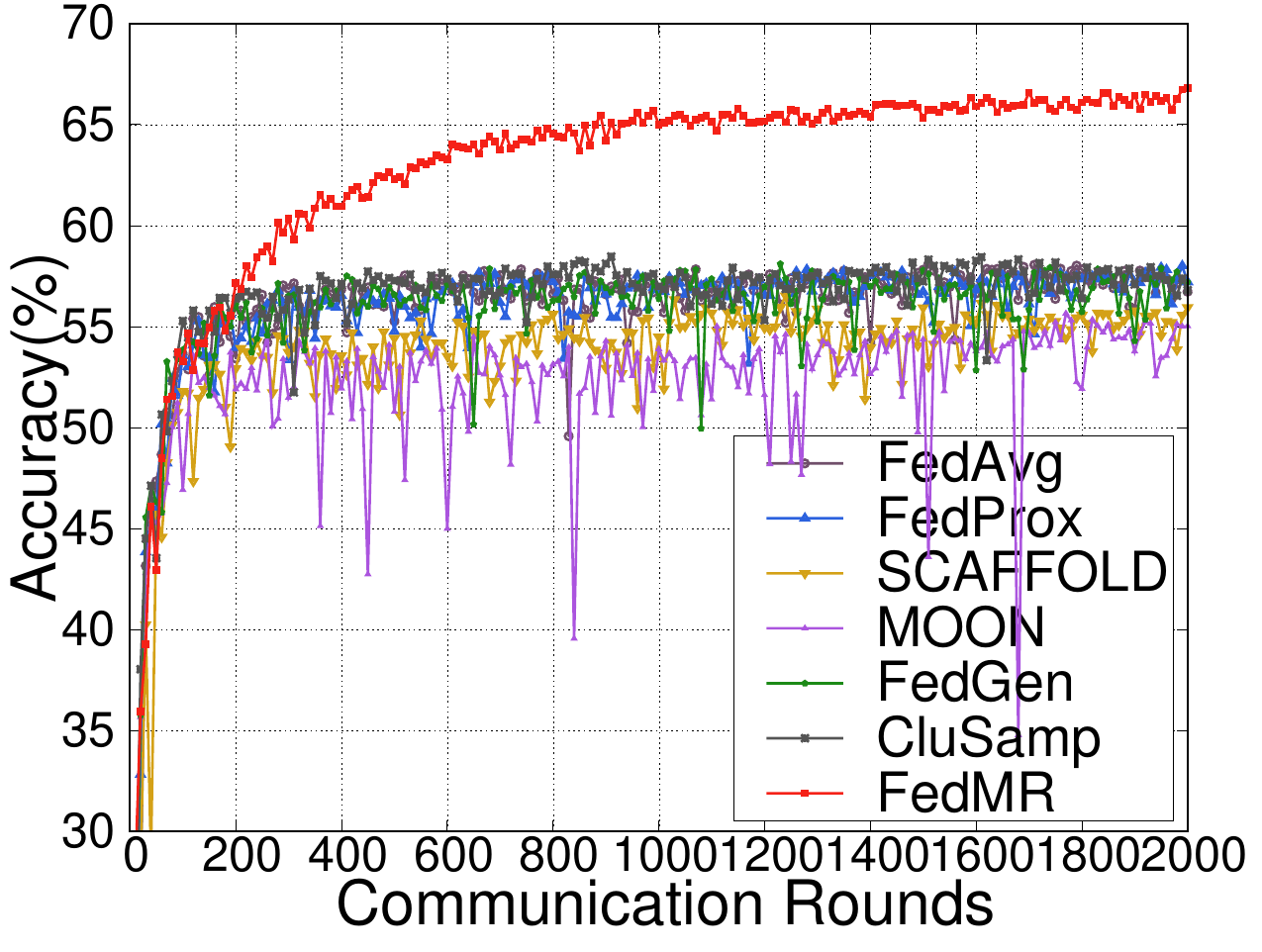}
	}
	\subfigure[CIFAR-100 with IID]{
		\centering
		\includegraphics[width=0.2\textwidth]{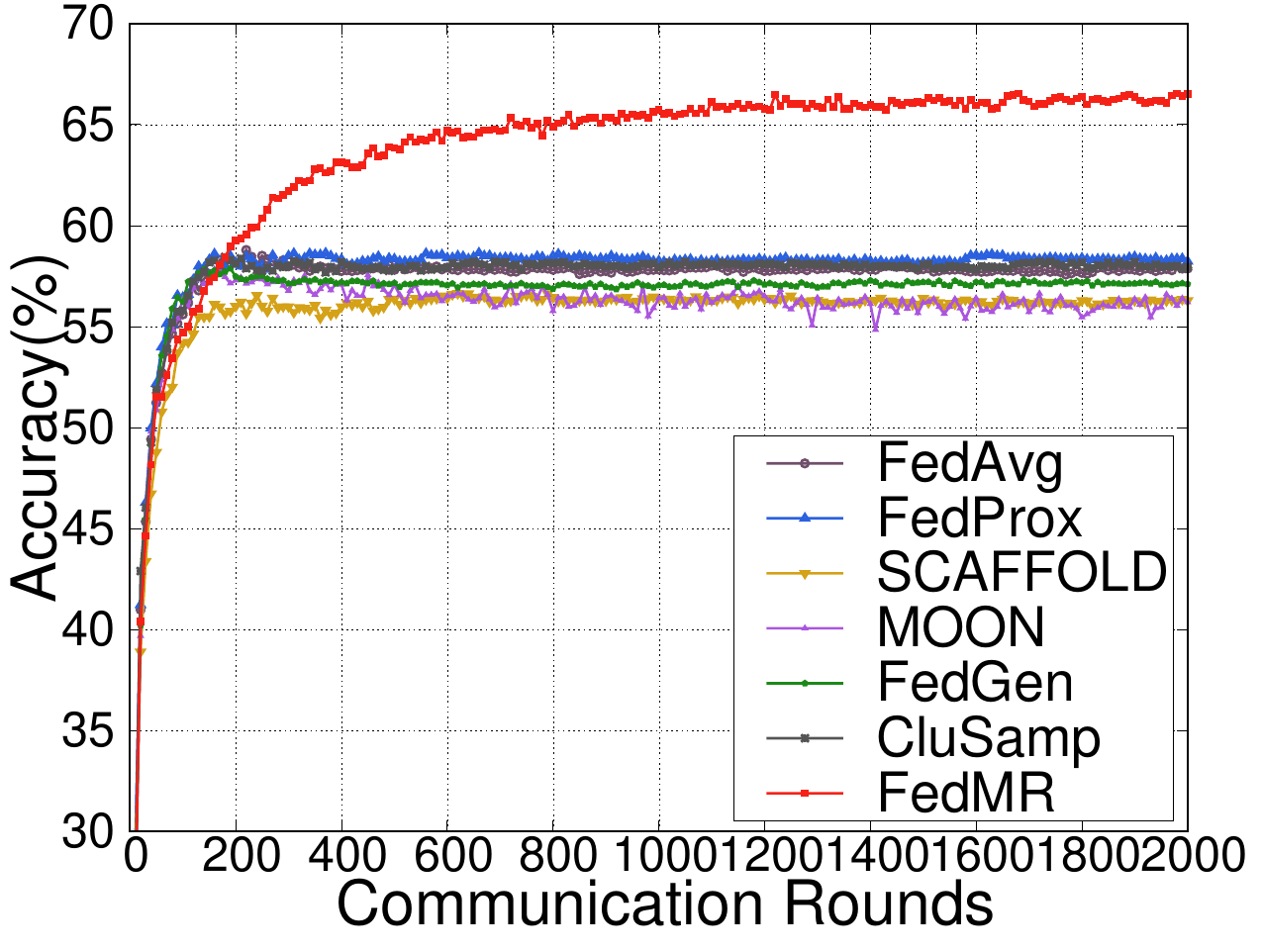}
	}
	\subfigure[FEMNIST]{
		\centering
		\includegraphics[width=0.2\textwidth]{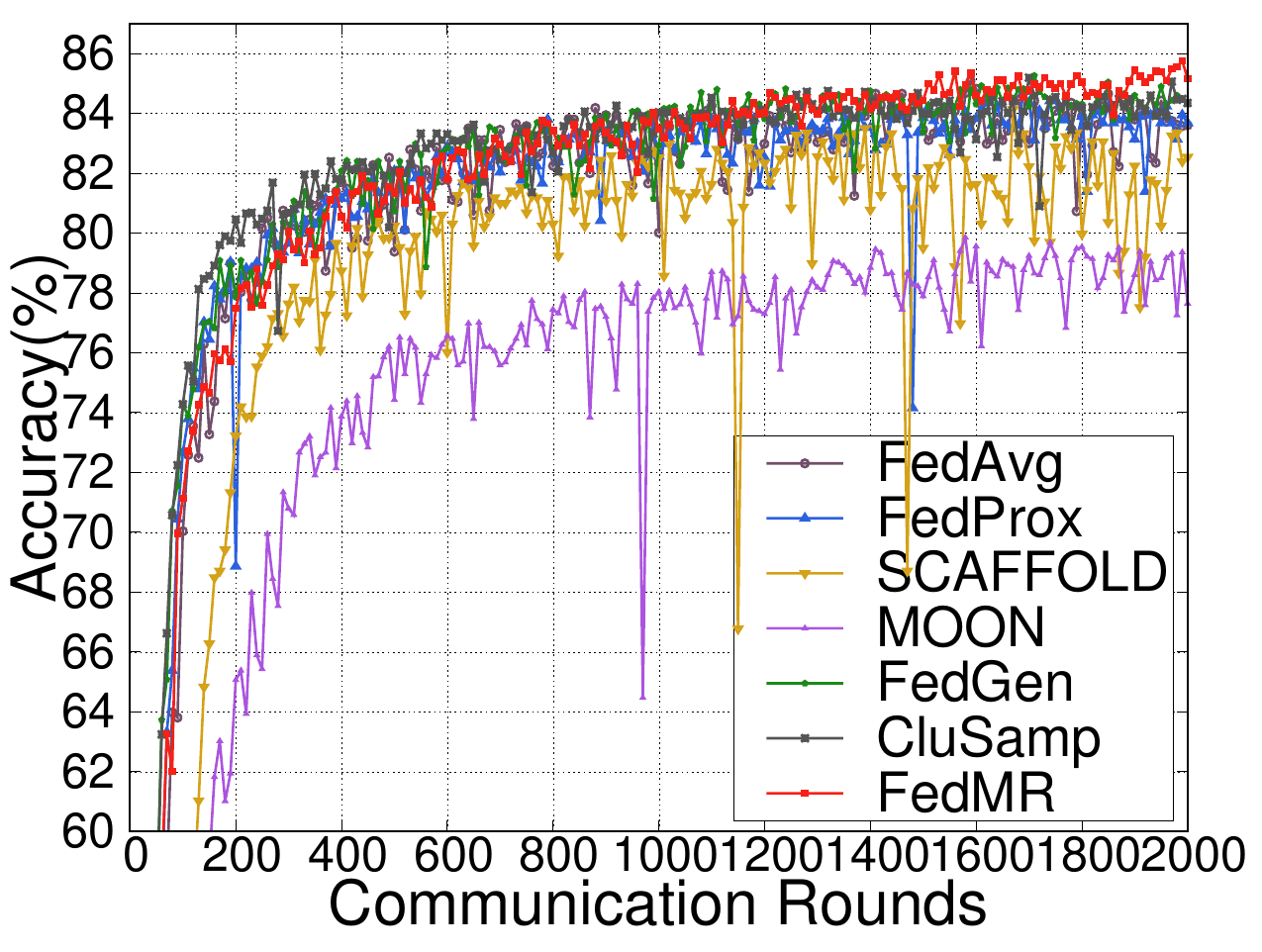}
	}
	\caption{Learning curves of different FL methods based on the VGG-16 model.}
	\label{fig:accuacy_vgg}
\end{figure}

In this section we presents all the experimental results.
Figure~\ref{fig:accuacy_cnn}-\ref{fig:accuacy_vgg} compare  the learning curves of FedMR with  all the six baselines based on the CNN, ResNet-20, and VGG-16, respectively.

\end{document}